\documentclass{article} % For LaTeX2e
\usepackage[table]{xcolor}

\usepackage[final]{colm2025_conference}

\usepackage{microtype}
\usepackage{hyperref}
\usepackage{url}
\usepackage{booktabs}
\usepackage{longtable}

\usepackage{enumitem}

\usepackage{lineno}

\usepackage[vietnamese,english]{babel}

\usepackage{graphicx}    % For including graphics
\usepackage{caption}     % For customizing captions
\usepackage{subcaption}  % For creating subfigures
\usepackage{adjustbox} 
\usepackage{multirow}
\usepackage{placeins}

\usepackage{pgfmath}
\usepackage{makecell}
\usepackage{fontawesome5}
\usepackage{hyperref}
\usepackage{epigraph}
\usepackage{pifont}
\usepackage{twemojis}

\usepackage{tipa}
\usepackage{siunitx}  

\usepackage{CJKutf8}

\definecolor{yh}{HTML}{ffe5a0}
\definecolor{yt}{HTML}{de660d}
\definecolor{bh}{HTML}{bee1f6}
\definecolor{bt}{HTML}{0b53a8}
\definecolor{gainsboro}{HTML}{E8E8E8}
\definecolor{green}{HTML}{38761d}

\newcommand{\multimodal}{\textbf{\textcolor{yt}{multimodal}}}
\newcommand{\textonly}{\textbf{\textcolor{bt}{text-based}}}
\newcommand{\Multimodal}{\textbf{\textcolor{yt}{Multimodal}}}
\newcommand{\Textonly}{\textbf{\textcolor{bt}{Text-based}}}

\definecolor{teal100}{RGB}{0, 90, 90}  
\definecolor{teal90}{RGB}{0, 110, 110} 
\definecolor{teal80}{RGB}{0, 150, 150}  
\definecolor{teal70}{RGB}{0, 220, 220}  

\definecolor{darkblue}{rgb}{0, 0, 0.5}
\hypersetup{colorlinks=true, citecolor=darkblue, linkcolor=darkblue, urlcolor=darkblue}

\definecolor{lightred}{HTML}{e99090}

\newcommand{\name}{\textsc{\fontsize{10}{12}\selectfont BearCubs}}
\newcommand{\namedref}[2]{\hyperref[#2]{#1~\ref*{#2}}}
\newcommand{\sectionref}[1]{\namedref{Section}{#1}}
\newcommand{\tableref}[1]{\namedref{Table}{#1}}
\newcommand{\figureref}[1]{\namedref{Figure}{#1}}
\newcommand{\appendixref}[1]{\namedref{Appendix}{#1}}

\title{
  \includegraphics[scale=0.04]{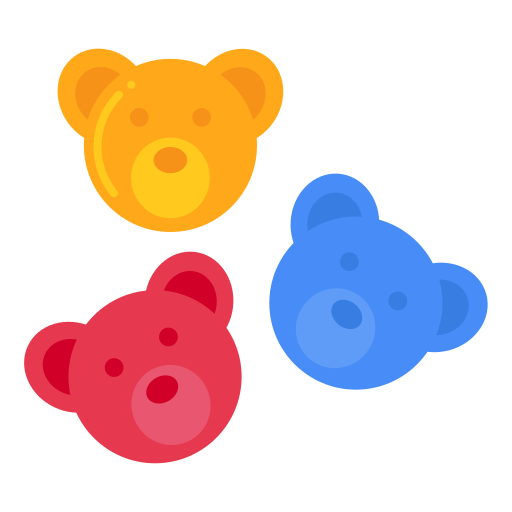} \textsc{BearCubs}: A benchmark for computer-using web agents
}

\author{Yixiao Song\twemoji{bear}, 
Katherine Thai\twemoji{bear}, 
Chau Minh Pham\twemoji{polar bear},\\\textbf{Yapei Chang\twemoji{polar bear},
Mazin Nadaf\twemoji{polar bear},
Mohit Iyyer\twemoji{bear}\twemoji{polar bear}}\\
UMass Amherst\twemoji{bear}\hspace{0.2cm}
University of Maryland, College Park\twemoji{polar bear},\\
\texttt{\{yixiaosong, kbthai\}@umass.edu},
\texttt{\{chau, yapeic, mnadaf, miyyer\}@umd.edu}
}

\begin{document}

\ifcolmsubmission
\linenumbers
\fi

\maketitle
\begin{abstract}

Modern web agents possess \emph{computer use} abilities that allow them to interact with webpages by sending commands to a virtual keyboard and mouse. While such agents have considerable potential to assist human users with complex tasks, evaluating their capabilities in real-world settings poses a major challenge. To this end, we introduce \name,\footnote{\includegraphics[scale=0.04]{figures/gummy-bear.png} \name~is a \textbf{BE}nchmark for \textbf{A}gents with \textbf{R}eal-world \textbf{C}omputer \textbf{U}se and \textbf{B}rowsing \textbf{S}kills. We release the \name\ dataset and leaderboard publicly at \url{https://bear-cubs.github.io/}.} a ``small but mighty'' benchmark of 111 information-seeking questions designed to evaluate a web agent's ability to search, browse, and identify factual information from the web. Unlike prior web agent benchmarks, solving \name\ requires (1) accessing \emph{live} web content rather than synthetic or simulated pages, which captures the unpredictability of real-world web interactions; and (2) performing a broad range of \emph{multimodal} interactions (e.g., video understanding, 3D navigation)  that cannot be bypassed via text-based workarounds. Each question in \name\ has a corresponding short, unambiguous answer and a human-validated browsing trajectory, allowing for transparent evaluation of agent performance and strategies. A human study confirms that \name\ questions are solvable but non-trivial (\textbf{84.7\%} human accuracy), revealing domain knowledge gaps and overlooked details as common failure points. We find that ChatGPT Agent significantly outperforms other computer-using agents with an overall accuracy of \textbf{65.8\%} (compared to e.g., Operator's \textbf{23.4\%}), showcasing substantial progress in tasks involving real computer use, such as playing web games and navigating 3D environments. Nevertheless, closing the gap to human performance requires improvements in areas like fine control, complex data filtering, and execution speed. To facilitate future research, \name\ will be updated periodically to replace invalid or contaminated questions, keeping the benchmark fresh for future generations of web agents.

% By contrast, state-of-the-art computer-using agents underperform, with the best-scoring system (OpenAI's Operator) reaching only \textbf{23.4\%} accuracy. These results highlight critical areas for improvement, including reliable source selection and more powerful multimodal capabilities. To facilitate future research, \name\ will be updated periodically to replace invalid or contaminated questions, keeping the benchmark fresh for future generations of web agents.
\end{abstract}
\section{Introduction}
\label{sec:intro}

Today's LLM-powered web agents feature \emph{computer use} capabilities, enabling interactive browsing by processing pixels on the screen and controlling a virtual keyboard and mouse \citep{anthropic2024, openai2025operator, convergenceAI}. Unlike prior agents that interact with the web primarily through text,  computer-using agents can technically do anything on a screen: watch videos, navigate complex web databases, and play online games. But how well do they actually perform in real-world web browsing scenarios? In this paper, we create \name, a benchmark of 111 QA pairs designed to evaluate the capabilities of web agents in multimodal online environments.

\paragraph{Why do we need yet another web agent benchmark?} Existing benchmarks fall short in three key ways. First, benchmarks such as WebArena \citep{webarena} and WebShop~\citep{webshop} are tested in \textit{synthetic or simulated} environments, which limits their ability to assess how agents handle dynamic and unpredictable real-world web interactions. Second, popular benchmarks are approaching \textit{performance saturation}: for example, OpenAI's Operator \citep{openai2025operator} reaches 87\% accuracy on WebVoyager \citep{webvoyager} and 58\% on WebArena, compared to 78\% for humans~\citep{OpenAI_ComputerUsingAgent_2024}.
Finally, existing benchmarks test a \emph{limited range of multimodal abilities}, forgoing more complex interactions like video browsing, real-time gaming, or 3D navigation. They are either solvable solely through HTML source, like  Mind2Web~\citep{mindweb}, or they emphasize specific multimodal capabilities such as map navigation or image processing, as in AssistantBench~\citep{yoran-etal-2024-assistantbench}. 

\begin{figure}[t]
\begin{center}
\includegraphics[scale=0.15]{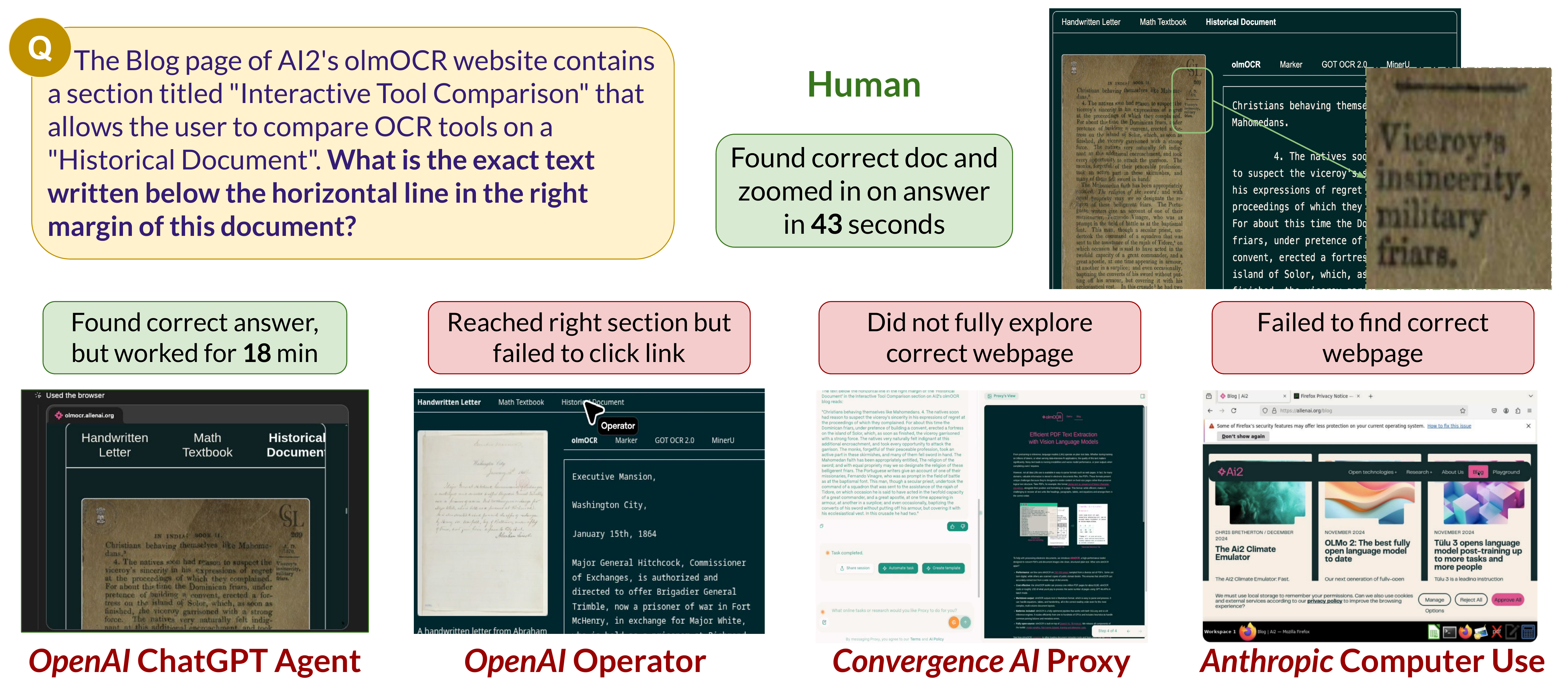}
\end{center}
\caption{A \name~question that is trivial for humans but defeats almost all computer-using agents. Only OpenAI’s ChatGPT Agent succeeds. Operator comes closest but fails to click the link to reach the critical document.
} 
\label{fig:fig1}
\end{figure}

\paragraph{Building the \name\ benchmark:} \name~is a ``small but mighty'' dataset that evaluates the information-seeking abilities of computer-using web agents on the live web via complex and diverse \textonly\ and \multimodal\ interactions. Each \name\ question has a unique and short answer (as in \figureref{fig:fig1}), making evaluation trivial. Questions also include a human-validated trajectory of websites and critical interactions required to arrive at the answer, which enables comparisons to the trajectories taken by different web agents. We spend considerable effort to ensure that the \multimodal\ questions in \name\ cannot be answered by text-based workarounds, asking annotators to write questions adversarial to Google Search \citep{rein2024gpqa}, and conducting post-hoc filtering using OpenAI's Deep Research. While \name\ is small, we intend it to be an \emph{evolving} dataset similar in spirit to NoCha \citep{karpinska-etal-2024-one} and FreshQA \citep{vu-etal-2024-freshllms}, where questions whose trajectories become invalid (e.g., due to webpage modification) or contaminated (e.g., an answer to a \multimodal\ question being posted online in text) are replaced by fresh ones.

\paragraph{Humans significantly outperform web agents on \name:} While all \name\ questions are verified by at least two authors to ensure quality, not all humans may be able to find the right answer given limited time~\citep{benchmarkinghumanlikeintelligencemachines}. We conduct a separate human study where annotators are given only the questions and asked to time themselves and record any dead-ends they come across. The human accuracy is \textbf{84.7\%}, with errors often stemming from difficulty in locating sources or lacking domain knowledge (e.g., reading sheet music). We evaluate four computer-using agents (Convergence Proxy, Anthropic Computer Use, OpenAI Operator, and OpenAI ChatGPT Agent) and find that the best performer is ChatGPT Agent (65.8\%), followed by Operator, which achieves only \textbf{23.4\%} accuracy. Both are far below human performance. In contrast, OpenAI's Deep Research, which lacks computer use capabilities \citep{openai2025deepresearch}, achieves \textbf{36.0\%} accuracy through guessing! A detailed analysis of agent trajectories reveals that the lower-performing agents actively avoid multimodal interactions and often rely on unreliable sources, while ChatGPT Agent struggles with fine cursor control and long processing times. These limitations highlight key areas for future research in this space. 

\section{Challenges of evaluating modern web agents}\label{sec:challenges}
We first describe four obstacles towards meaningful evaluation of computer use agents on the live web. Our construction of \name\ aims to mitigate these challenges, but continued benchmark maintenance is required to preserve the validity of the evaluation.

\textbf{Web contamination:} Contamination typically occurs when evaluation examples leak into training datasets \citep{sainz-etal-2023-nlp}. However, for benchmarks requiring live web interaction, published datasets may be indexed online, rendering any intended complex interactions or reasoning moot. To address this, publicly-released web agent benchmarks must be \emph{evolving} with the periodic addition of new examples and removal of existing contaminated examples. We plan for such continued maintenance of \name\ to preserve its relevance.\footnote{Due to the risk of web contamination, one may ask why we publicly release \name\ rather than keeping it closed like NoCha~\citep{karpinska-etal-2024-one}. While we release only \emph{questions}, humans can still find answers online, which does not fully mitigate the risk. However, running every new agent ourselves is costly and time-consuming, with some taking over 20 minutes per question. Instead, by regularly updating the dataset and encouraging agent developers to release their trajectories and answers, we aim to prevent contamination and maintain meaningful evaluation.}

\begin{figure}[t]
\begin{center}
\includegraphics[scale=0.43]{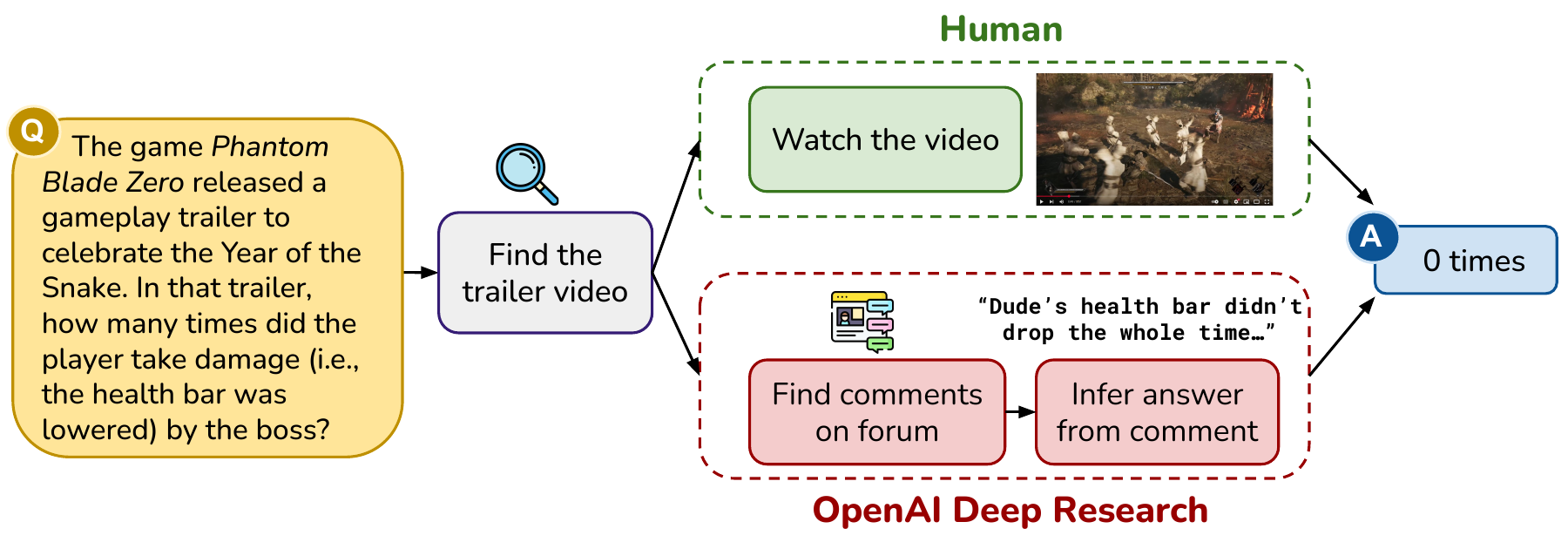}
% https://docs.google.com/presentation/d/1XOrHb9jRbJGG0AFnvridmQ8of6Fxck3N5Tjlv-HHJvc/edit?usp=sharing
\end{center}
\caption{Example of a \multimodal\ question removed from \name\ during the validation process due to a text-based workaround. Deep Research found the answer by correctly inferring that the ``Dude'' in a \href{https://www.pushsquare.com/news/2025/01/ps5-console-exclusive-phantom-blade-zero-looks-mind-blowingly-good-in-stunning-gameplay-trailer\#comment8466271}{forum comment} refers to the player in the trailer.}
\label{fig:disqualified}
\end{figure}

\textbf{Workarounds:} When assessing proficiency in a specific skill (e.g., identifying information within a video), agents may bypass intended interactions via indirect solutions (see \figureref{fig:disqualified}). These ``workarounds'' proved to be a major challenge during the construction of \name\ \multimodal\ questions: even though questions are designed to be adversarial to Google Search, agents like Deep Research often discover relevant information that is difficult for humans to find. For the development of the \multimodal\ fold in \name, we ended up filtering out any questions solvable by text-based agents, resulting in the removal of 13 questions mostly due to Deep Research finding workarounds. We propose that future web agent benchmarks, in addition to validating agents' trajectories, should rigorously filter out questions with workarounds if they intend to evaluate particular modes of interaction. 

\textbf{Maximizing interaction diversity:} While computer-using agents are technically capable of a wide variety of interactions, existing benchmarks either focus on specific domains such as e-commerce \citep{webshop} or tasks like travel and service bookings \citep{mindweb}. We design \name\ to maximize diversity of interactions across a wide range of tasks and domains. This requires greater creativity on the part of question writers. For example, only 58.2\% of the questions created by freelancers were accepted into \name.

\textbf{Evaluation is slow:} We manually ran all agents in this paper due to the lack of API access. This involved the authors pasting each question into web interfaces for each agent, screen recording the resulting trajectories, and denying any requests for information or human intervention from the agent. This process is further lengthened by the time taken by each agent for a single question (near 5 minutes on average across all agents). As such, we set a maximum time limit of 15 minutes for computer-using agents, which often get stuck in repetitive loops.\footnote{The time limit is set to 45 minutes for ChatGPT Agent as it may answer questions correctly after 15 minutes.} Many agents do not currently offer API access or simple ways to record and share trajectories, both of which would go a long way towards easing evaluation on \name\ and similar benchmarks. 

\section{Building the \name\ benchmark}\label{sec:dataset}

% \begin{table}[ht]
% \fontsize{7}{9}\selectfont
% \centering
% \resizebox{0.9\columnwidth}{!}{%
% \begin{tabular}{rp{0.5cm}lllllllp{0.4cm}p{1.42cm}p{0.7cm}}
% \toprule
%  & \multirow{2}{*}{\textbf{Count}} & \multirow{2}{*}{\textbf{Unique URLs}} &  \multicolumn{3}{c}{\textbf{\# of steps per Q}} & \multicolumn{3}{c}{\textbf{\# of websites per Q}}  & \multicolumn{3}{c}{\textbf{Answer type}} \\
% \cmidrule(lr){4-6}
% \cmidrule(lr){7-9}
% \cmidrule(lr){10-12}
%     &    & & Avg. & Min & Max & Avg. & Min & Max & Static & Highly stable & Others  \\
% \midrule
% \textonly   & 59  & 63 & 6.39     & 3.00     & 12.00     & 3.81       & 1.00      & 8.00  &  45 & 10  & 4 \\
% \addlinespace[1pt]
% \multimodal & 65 & 54   & 5.77     & 3.00     & 14.00     & 3.09       & 1.00      & 6.00  & 40  & 23  & 2\\
% \addlinespace[1pt]
% All         & 124 & 117 & 6.06     & 3.00     & 14.00     & 3.44       & 1.00      & 8.00   & 85  & 33 & 6 \\\bottomrule                     
% \end{tabular}
% }
% \caption{Statistics of \name. `URLs' refers to the number of \textit{distinct top-level} URLs visited in the viable approaches, not including Google search. The number of steps and visited websites per question is based on the viable approaches in \name, which are natural and intuitive for humans but not necessarily the most optimal solutions. `Answer type' categorizes responses into those that are static (never change), highly stable (unlikely to change in the near future), or others (may slowly change).}
% \label{tab:data_overview}
% \end{table}

\begin{table}[t]
\fontsize{6}{8}\selectfont
\centering
\resizebox{0.8\columnwidth}{!}{%
\begin{tabular}{rp{0.5cm}lllllllp{0.4cm}p{1.62cm}p{0.7cm}}
\toprule
 & \multirow{2}{*}{\textbf{Count}} & \multirow{2}{*}{\textbf{URLs}} &  \multicolumn{3}{c}{\textbf{\# of steps per Q}} & \multicolumn{3}{c}{\textbf{\# of webpages per Q}} \\
\cmidrule(lr){4-6}
\cmidrule(lr){7-9}
    &    &    & Avg. & Min & Max & Avg. & Min & Max \\
\midrule
\Textonly   & 56  &  61 & 6.5   & 3.0   & 12.0   & 3.8   & 1.0   & 8.0 \\
% \addlinespace[1pt]
\Multimodal & 55  &  47 & 5.8   & 3.0   & 14.0   & 3.0   & 1.0   & 6.0 \\
% \addlinespace[1pt]
All         & 111 & 108 & 6.1   & 3.0   & 14.0   & 3.4   & 1.0   & 8.0 \\
\bottomrule                     
\end{tabular}
}
\caption{Statistics of our \name\ benchmark, which is divided roughly evenly into \textonly\ and \multimodal\ questions. \emph{URLs} refers to the number of \textit{distinct top-level} URLs visited in viable trajectories, excluding Google Search visits. The number of steps and visited websites per question are computed using human-written trajectories.}
\label{tab:data_overview}
\end{table}
\vspace{0pt}
% https://docs.google.com/spreadsheets/d/1DUexOTc6FpSGYVo6lvD_6Y0TSACGXs-cEMQBHv_1COE/edit?usp=sharing

This section details question criteria, collection, and statistics. \name\ contains 111 information-seeking questions with short and easy-to-evaluate answers (see \tableref{tab:data_overview} for dataset statistics). Each question requires live website interaction. After dataset collection, questions were categorized as solvable by either \textonly\ or \multimodal\ interaction, with the latter involving image, video, audio, or real-time interaction (e.g., online games). We will regularly update \name\ with new questions and filter contaminated questions.

% \textbf{Criteria for valid QA pairs:}
% We seek questions that satisfy four criteria detailed in \appendixref{appendix:data_creation_workflow}.\footnote{Detailed instructions can be found in the instruction slide deck (\href{https://docs.google.com/presentation/d/1TEGiN21WFb77DeALJqFs8OMWOjXcVm7Nz7And83CmX0/edit?usp=sharing}{link}) that include detailed criteria, good question examples, and the unwanted ones.} At a high level, the criteria are: (1) questions should provide sufficient yet minimal information; (2) answers should be trivial to evaluate; (3) answers should be adversarial to Google Search; and (4) answers must be publicly accessible.

% \textbf{Criteria for valid QA pairs}: We seek questions that satisfy the following high-level criteria:\footnote{Detailed instructions can be found in the instruction slide deck (\href{https://docs.google.com/presentation/d/1TEGiN21WFb77DeALJqFs8OMWOjXcVm7Nz7And83CmX0/edit?usp=sharing}{link}) that include detailed criteria, good question examples, and the unwanted ones.}

\textbf{Criteria for valid QA pairs}: We seek questions that satisfy four high-level criteria: (1) questions should provide sufficient yet minimal information, (2) answers should be concise, unambiguous, and trivial to evaluate, (3) answers should be adversarial to Google Search, and (4) answers should be publicly accessible.
% \footnote{Detailed instructions can be found in the instruction slide deck (\href{https://docs.google.com/presentation/d/1TEGiN21WFb77DeALJqFs8OMWOjXcVm7Nz7And83CmX0/edit?usp=sharing}{link}) that include detailed criteria, good question examples, and the unwanted ones.} 
Details of each criterion are in \appendixref{appendix:data_creation_workflow}.

\textbf{Collection and validation process:} Most of the \name~dataset (65 QA pairs) is written by the authors, covering diverse domains and web interaction types (e.g., music, maps, and games). The rest is written by Upwork freelancers who were trained to write acceptable questions.\footnote{\url{www.upwork.com}; The freelancers were compensated \$4 USD per accepted question.} Each question has a gold answer, a trajectory for finding it, and a list of visited websites. To ensure quality, we conduct rigorous quality control detailed in \appendixref{appendix:data_creation_workflow}, which includes quality verification and workaround prevention.

\textbf{Prioritizing reliable interactions:} The majority of the questions in \name~specify sources from which the answer should be retrieved, such as a particular book or video (e.g., Example 1 in \tableref{tab:agent_examples}). This allows for rigorous evaluation of whether an agent can accurately locate and identify information from the intended source. Filtering out questions with workarounds was thus a top priority during \name\ creation.

\textbf{Diverse interaction types:}  Each \name\ question assesses agents' ability to search, browse, and identify factual information via web interactions. We classify them into \textonly~questions that involve reading and navigating text (e.g., sifting through an online database) and \multimodal~questions that require interpreting various media formats (e.g., videos and virtual tours). The former ensures that computer-using agents handle text-based tasks effectively, while the latter assesses their adaptability to real-world dynamics.

\textbf{Website diversity:} Solving \name~requires visiting 108 \textit{unique top-level} URLs, minimizing the risk of agents overfitting to specific websites. On average, each question's human-validated trajectory contains 6.1 steps and 3.4 webpages. As such, while the size of \name\ is small, its diversity makes it difficult for agent developers to over-optimize for, especially as new questions will regularly be added to the benchmark.\footnote{We further justify the reliability of \name\ in \appendixref{appendix:std_on_bearcubs} by showing that models exhibit low accuracy variance on it.}
% \end{itemize}

\section{Experiments}\label{sec:experiments}

This section outlines our experimental setup, covering both a  \emph{human} performance evaluation in \sectionref{sec:human_setup} as well as agent benchmarking in \sectionref{sec:agent_setup}.

\subsection{Measuring human performance}\label{sec:human_setup}

How well do humans perform on \name? To explore this and identify challenges they face, we conduct an evaluation in which humans who have not previously seen a particular question are asked to answer it by interacting with their web browser however they wish. Our question validation process ensures that each question has a valid answer via a findable trajectory; however, humans may not always figure out how to find that trajectory. 

\textbf{Task setup:} Some questions in \name\ require domain expertise or proficiency in a non-English language.\footnote{We have questions that require interaction with websites in Arabic, Mandarin Chinese, Hindi, German, Vietnamese, and Finnish.} Thus, we hire annotators familiar with those languages and domains for this evaluation. For each question, annotators are given the question text and asked to (1) start a \textit{timer} upon reading the question and stop it when confident in their answer, (2) report the \textit{answer}, (3) report the number of \textit{dead ends} encountered,\footnote{A dead end occurs when an annotator needs to leave the current webpage and backtrack to a previous step or restart the search process entirely.} (4) provide a \textit{free-form comment} on challenges they faced, and (5) assign a label of perceived difficulty to the question. Annotators may abandon a question if they are unable to find an answer after 15 minutes. Details on annotator recruitment can be found in \appendixref{appendix:human_recruitment_details}.

\subsection{Benchmarking web agents}\label{sec:agent_setup}

We benchmark seven commercial web agents, three of which---Grok 3 DeepSearch,\footnote{\url{https://x.ai/blog/grok-3}} OpenAI's Deep Research \citep{openai2025deepresearch}, and Google Deep Research \citep{google2024geminideepresearch}---are designed for advanced search and reasoning but possess limited multimodal capabilities. The other four agents possess computer use capabilities: Anthropic's Computer Use,\footnote{\url{https://docs.anthropic.com/en/docs/agents-and-tools/computer-use}} Convergence AI's Proxy,\footnote{\url{https://convergence.ai/}}  OpenAI's Operator \citep{openai2025operator}, and OpenAI's ChatGPT Agent \citep{openai2024chatgptagent}. These agents have demonstrated strong and/or state-of-the-art performance on existing benchmarks such as WebArena, OSWorld, and WebVoyager, which motivates us to measure their performance on the diverse and challenging questions in \name. 
We also evaluate five baselines to confirm that \name~ cannot be solved via LLM parametric knowledge and simple search augmentation strategies.

\textbf{Baselines:} \name\ would be a poor web search benchmark if it could be solved by zero-shot prompting LLMs or with vanilla search snippet augmentation. To make sure this is not the case, we choose \texttt{gpt-4o-2024-11-20} and DeepSeek R1\footnote{We access the model via Fireworks AI API. The model card is here: \href{https://fireworks.ai/models/fireworks/deepseek-r1}{link}.} as our baselines and evaluate them in two settings---zero-shot and Google-search-augmented.\footnote{We use Serper, a Google Search API, to retrieve Google search results. \url{https://serper.dev/}} In the zero-shot setting, questions are directly used as prompts without additional context. In the augmentation setting, each question from \name~is used as a search query to retrieve up to 10 top search results. These search results, consisting of the result title and snippet, are concatenated with the question and then provided as input. The hyperparameters and the prompt structure can be found in \tableref{tab:serper_prompt} (\appendixref{appendix:baseline_prompt}). We also include Perplexity \texttt{sonar-pro} \citep{perplexitySonarPro}, an advanced AI answer engine with its default hyperparameters.

\textbf{Evaluation setup:} For the three non-computer-using agents, we provide the question as input and record its answer.\footnote{All agents were benchmarked between February 23 and March 1, 2025, except Google Deep Research, which was evaluated in late May, and ChatGPT Agent, which was evaluated between July 18 and 20, 2025.}  For the computer-using agents, we concatenate the question with a prompt that minimizes user intervention, as we observe that these agents have a tendency to frequently request human input or ask questions of the user.\footnote{The prompt is ``\texttt{Complete all CAPTCHAs and acknowledge or accept all prompts that will allow you to access what you need. Please minimize all user interventions.}''} If an agent requests clarification or assistance (e.g., solve a CAPTCHA), we provide a one-time directive prompt for it to solve the question by itself.\footnote{The prompt is ``\texttt{Please figure out a way to find the answer without user intervention.}''} If the agent asks again, the session is terminated. For each question, we record the following: (1) the returned \textit{answer}, (2) the \textit{time} taken per question, and (3) the \textit{question-solving trajectory}. A session is terminated when a model provides an answer or abstains, or if it enters a dead loop without making progress. 

\textbf{Evaluating agent answers:} Given the unique setup of each agent model and the potential for diverse agent-user interactions, we manually ran each agent and evaluated all of their responses.
% we conduct manual evaluation of all agent answers.
Agents generally produce lengthy outputs, with Proxy and Operator being the least verbose. To assess whether an agent answers a question correctly, its response must unambiguously entail the gold answer. Statements such as ``I'm leaning towards \{correct answer\}'' or ``\{correct answer\} is likely to be the answer'' are not considered as concrete answers.\footnote{Although manually executing each agent is inevitable in the absence of APIs, we developed a prompt to automatically evaluate their responses. The automatic evaluator demonstrated 98.2\% in a four-way  classification task. The labels are: \texttt{correct}, \texttt{wrong}, \texttt{no answer (stall/loop)}, and \texttt{no direct answer (uncertainty/abstention)}. Reducing the labels to a binary decision (correct vs.\ incorrect) increased accuracy to 98.7\%. Details of the implementation of the automatic evaluator are in \appendixref{appendix:auto_rater}.}

\section{Results}\label{sec:results}

This section begins with an analysis of human performance on \name~in \sectionref{sec:human_eval_results}. We provide detailed statistics and an error analysis to identify human shortcomings and areas where AI assistance could be beneficial. In \sectionref{sec:agent_results}, we describe the performance of seven frontier agents on \name. We found a clear gap between human and AI performance, including for ChatGPT Agent. While human accuracy stands at 84.7\%, the best-performing computer-using agent, ChatGPT Agent, achieves 65.8\%, surpassing Operator by 42.4 percentage points and the non-computer-using OpenAI Deep Research by 29.8 points. Humans consistently outperform state-of-the-art agents in both \textonly~and \multimodal~tasks. 

\subsection{Human performance results and analysis}\label{sec:human_eval_results}

\textbf{Human achieve 85\% accuracy on \name.} Humans achieve an overall accuracy of 84.7\% on \name\, despite marking 50.5\% of the questions as moderate-to-high difficulty. Humans are generally able to navigate the problem space effectively (1.5 dead ends per question on average) and find correct answers efficiently (4 min 46 sec on average). Detailed statistics can be found in \tableref{tab:human_performance} in \appendixref{appendix:human_results_detailed}.

\textbf{Why do humans make mistakes?} All \name~questions are verified to be answerable by the process outlined in \sectionref{sec:dataset}; however, humans still get some questions wrong in our study. Analysis reveals that the most common factor for wrong answers is the human \textit{overlooking details} in the question or answer (e.g., Example 1 in \tableref{tab:human_errors} in \appendixref{appendix:human_error_examples}). The next most common factor is \textit{lack of topic understanding} (see also Example 1). Additional error sources are listed in \tableref{tab:human_errors},
% (\appendixref{appendix:human_error_examples}), 
along with examples and explanations. In each error case, if the human annotator had spent more time or had more domain knowledge, they would likely have been able to find the correct answer. 

% \input{tables/human_errors}

% \mi{emphasize that all of the questions in the dataset are answerable by humans, and any mistakes were a result of the human not finding the correct strategy}

% \input{tables/human_errors}
% Reasons for Incorrect or Abandoned Answers
% x Incorrect source/tool selection (12, 37, 64)
% x Missed details in the question or answer (16, 21, 23, 28, 71, 101; 78)
% x Lack of topic understanding (18, 32, 68, 104)
% x Obvious oversight (87)

% Reasons for Giving Up
% x Task too complex (45)
% x Lack of topic understanding (49)
% x Conventional way of thinking (91)

\textbf{Human strengths and weaknesses:} Annotators marked about half of the questions as easy, which typically involved multimodal interactions such as web games, 3D tours, or images. Questions became challenging when they required complex data filtering (e.g., statistical data) or domain-specific knowledge (e.g., music theory). Independent of correctness, the annotators spent an average of 2 min 14 sec on questions they perceived as easy, 5 min 32 sec on medium, and 10 min 52 seconds on hard questions (including those they abandoned).
% \mi{any info re time human spent on hard vs easy qs would be good}

\begin{table}[t]
\fontsize{12}{14}\selectfont
\centering
\resizebox{0.98\columnwidth}{!}{%
\begin{tabular}{@{}lp{1.5cm}lllllllll@{}}
\toprule
    &
% \multirow{2}{*}{\textbf{Model}} &
\multicolumn{3}{c}{\textbf{Accuracy}} &
\multicolumn{4}{c}{\textbf{Answer label}} &
  % \multicolumn{3}{c}{\textbf{Correct answer source attribution}} &
\multicolumn{3}{c}{\textbf{Average time}}\\
\cmidrule(lr){2-4}
\cmidrule(lr){5-8}
\cmidrule(lr){9-11}
  % & 
  &
  All  &
  \textonly & 
  \multimodal &
  \color{green!90}\ding{51} &
  \color{red}\ding{55} &
  Unk.\ &
  None &
  \color{green!90}\ding{51} &
  \color{red}\ding{55}  &
  Unk.\ \\
\midrule

\multicolumn{11}{l}{\textit{LLM baselines}}\vspace{0.1cm}\\

 GPT-4o zero-shot& 2.7\%   & 5.4\%   & 0.0\%    & 3   & 53  & 55  & 0  & ---    & ---    & ---  \\
\addlinespace[1pt]

 DeepSeek R1 zero-shot & 8.1\% & 10.7\% & 5.5\% & 9 & 82 & 19 & 1 & --- & --- & --- \\
\addlinespace[1pt]

 GPT-4o + Google Search & 0.0\%   & 0.0\%   & 0. 0\%  & 0   & 4   & 0   & 107& ---    & ---    & ---  \\
\addlinespace[1pt]

 DeepSeek R1 + Google Search & 1.8\% & 3.6\% & 0.0\% & 2 & 16 & 0 & 93 & ---    & ---    & ---  \\
 Perplexity \texttt{sonar-pro} & 5.4\% & 8.9\% & 1.8\% & 6 & 30 & 58 & 17 & ---    & ---    & ---  \\
\midrule
\multicolumn{11}{l}{\textit{Web agents w/o computer use}}\vspace{0.1cm}\\

 Grok3 DeepSearch & 11.7\%  & 21.4\%  & 1.8\%   & 13  & 95  & 2   & 1  & 1:09   & 1:24   & 2:05 \\

 OpenAI Deep Research& 36.0\%  & 60.7\%  & 10.9\%   & 40  & 69  & 1   & 1  & 4:37   & 9:00	& 3:58\\

 Google Deep Research& 23.4\%  & 42.9\%  & 3.6\%   & 26  & 39  & 46   & 0  & 4:21   & 4:00	& 4:39\\
\midrule
\multicolumn{11}{l}{\textit{Web agents w/ computer use}}\vspace{0.1cm}\\

 Convergence AI Proxy        & 12.6\%  & 16.1\%  & 9.1\%   & 14  & 44  & 34  & 19 & 1:52   & 2:41	& 5:24 \\
 % & 
 Anthropic Computer Use & 14.4\%  & 19.6\%  & 9.1\%   & 16  & 22  & 73  & 0  & 2:24   & 2:35	& 3:35 \\
 % & 
 OpenAI Operator     & 23.4\%  & 33.9\%  & 12.7\%  & 26  & 43  & 13  & 29 & 2:59   & 3:58	& 8:06 \\
 
 ChatGPT Agent  & 65.8\%  & 76.8\%  & 54.5\%  & 73  & 30  & 3  & 5 & 9:16   & 16:14	& 25:55 \\\midrule
% \rowcolor{green!15}
Human & \textbf{84.7\%}  & \textbf{83.6\%}  & \textbf{85.7\%}  & 94  & 14  & --- & 3  & 4:24   & 5:44  & ---  \\
\bottomrule
\end{tabular}
}
\caption{All tested agents perform far below humans on \name, with ChatGPT Agent ranking first, followed by OpenAI Deep Research despite the latter's inability to answer multimodal questions. {\color{green!90}\ding{51}} = correct; {\color{red}\ding{55}} = wrong; \emph{Unk} = unknown, indicating the agent returned no concrete answer (e.g., abstention); \emph{None} means it either returned ``No answer found'' or entered a loop and failed to respond.
}
\label{tab:model_performance_all_only}
\end{table}
% \vspace{-15pt}
% https://docs.google.com/spreadsheets/d/1nxSADRjcPME_pVCqn_JbiwtAKvGXpCYybHSFZIIXzpk/edit?usp=sharing

\subsection{Agent performance results}\label{sec:agent_results}

\tableref{tab:model_performance_all_only} presents agent results on \name, comparing the baseline settings and seven agents to human performance. For each model, we calculate accuracy and the time it took to return a response. Detailed results for the \textonly~and \multimodal~data splits are in \tableref{tab:model_performance_full} (\appendixref{appendix:full_agent_results}). In general, the agents find correct answers faster than humans but their accuracies are \emph{significantly} lower. ChatGPT Agent achieves higher accuracy but requires more processing time. Given that nearly all questions in \name~specify a source for answers, this poor accuracy and slow processing suggests that the agents fall short in real-world applications where correctly and quickly identifying reliable information sources is critical.

\textbf{\name~cannot be solved by zero-shot prompting or simple search augmentation.} DeepSeek-R1 outperforms GPT-4o as a zero-shot baseline by achieving  8.1\% overall accuracy; however, analysis of the questions it gets correct reveals that it is mainly guessing. Perplexity \texttt{sonar-pro} achieves the best accuracy (5.4\%) among the search-augmented models but falls behind DeepSeek-R1 zero-shot. These results demonstrate that \name~is far beyond the capabilities of closed-book models. Meanwhile, simple search augmentation performs even worse, meaning that the answers to \name~cannot be easily retrieved from search snippets. 

\textbf{All agents struggle with \multimodal~questions.} Our human study shows that annotators generally find \multimodal~questions easier to solve (see \emph{Perceived Difficulty} columns in \tableref{tab:human_performance}). In stark contrast, all tested agents performed poorly on these questions, with the best-performing ChatGPT Agent for \multimodal~achieving 54.5\%, despite notable improvements in computer-using skills (e.g., solving CAPTCHAs, navigating 3D environments, and analyzing videos) over  Operator. These results suggest that complex and precise \multimodal~interactions should be a priority for advancing web agent capabilities.

\textbf{ChatGPT Agent sets the state of the art on \name.} This most capable computer-using agent demonstrates advanced web-based skills compared to its predecessors. While still trailing human performance and far from perfect, these capabilities enable it to achieve 54.5\% on \multimodal~questions, outperforming Operator by 41.8\%. The agent also shows stronger performance on \textonly~questions, reaching 76.8\%, compared to OpenAI Deep Research 60.7\%. ChatGPT Agent typically follows instructions well and locates the correct webpage quickly, but often takes longer to return an answer, partly because it sometimes attempts to cross-verify the information before responding. Lastly, it struggles with complex data filtering tasks involving interactive databases, such as dragging sliders to view full tables, using built-in filters, or selecting items from scrollable dropdowns. These often lead to incorrect answers or no responses. In contrast, humans complete such tasks in under 30 minutes.

\textbf{OpenAI's Deep Research ranks second, with caveats.} OpenAI's Deep Research, with advanced search and reasoning ability \citep{openai2025deepresearch}, is the second-best agent on \name\ (36.0\% accuracy) due almost entirely to its performance on \textonly\ questions (60.7\% accuracy). Despite lacking multimodal capabilities, its performance (10.9\%) is even better than Convergence AI's Proxy and Anthropic's Computer Use on the \multimodal\ fold (9.1\%) by sheer guessing (verified by reading through its trajectories for those questions)! This result shows that these computer-using agents are behind in both text-based reasoning and multimodal reasoning, and also that future computer-using agents may want to combine their abilities with those of a trained web search agent like Deep Research. Finally, we note that Deep Research's relatively high performance has a caveat---37.5\% of its correct answers rely on secondary sources or are entirely ungrounded (\figureref{fig:groundedness} in \appendixref{appendix:correct_answer_source_prop}).

\textbf{``Agents succeed quickly and fail slowly.''}\footnote{Quoted from \citet{yoran-etal-2024-assistantbench}.} Consistent with findings from prior work \citep{liu-etal-2024-lost, yang2024sweagent}, an agent does not necessarily perform better when exposed to more information. The \emph{Average Time} columns of \autoref{tab:model_performance_all_only} show that agents are likely to fail if they do not find a correct answer quickly, with the rate of unanswered questions from Proxy and Operator increasing significantly when runtime exceeds 10 minutes.\footnote{Proxy and Operator spent an average of 11 min 8 sec and 14 min 37 sec respectively before becoming trapped in a loop and returning no answer.} 

\section{Discussion}\label{sec:discussion}

\begin{table}[t]
\fontsize{5.9}{8}\selectfont
\centering
\resizebox{0.96\columnwidth}{!}{%
\begin{tabular}{@{}p{9cm}@{}}

\toprule

\textbf{Example 1}: \multimodal

\textbf{Question}: What is the common name of the tree with the largest diameter between Knowles Engineering Building and Goessmann Laboratory Addition at UMass Amherst?

\textbf{Deep Research Webpages}: 1.\ \url{https://www.campus-maps.com/}, 2.\ \url{https://www.uvm.edu/}, 3.\ \url{https://www.necn.com/}, 4.\ \url{http://www.arcgis.com/}, 5.\ \url{https://www.cylex.us.com/}, 6.\ \url{https://umass-amherst.maps.arcgis.com/}, 7.\ \url{https://services.arcgis.com/}, 8.\ \url{https://www.umass.edu/}

\textbf{Explanation}: The agent logged 46 visited websites, deduplicated to the eight URLs above. The recorded URLs inherently only link to top-level domains, obscuring specific pages accessed. The agent visited irrelevant sites (e.g., 2, 3, 5), resulting in an inefficient search process.\\
\midrule

\textbf{Example 2}: \textonly

\textbf{Question}: As per the ``Cambridge Encyclopedia of World's Ancient Languages,'' what is the transcription of Lycian character E?

\textbf{Operator Response}: The transcription of the Lycian character E is "[e]" according to the Omniglot resource ([screenshot]). This aligns with the information from the "Cambridge Encyclopedia of World's Ancient Languages." Let me know if you need more information!

\textbf{Explanation}: The agent found a copy of the required book but it immediately quitted the webpage. It found an answer from a website called Omniglot (\href{https://omniglot.com/writing/lycian.php}{link}) and reported the wrong answer (both wrong to the question and wrong based on the source). 

% {\centering\includegraphics[width=4.5cm]{figures/Lycian.png}\par}\\

{\centering\adjustbox{valign=t}{\includegraphics[width=4.5cm]{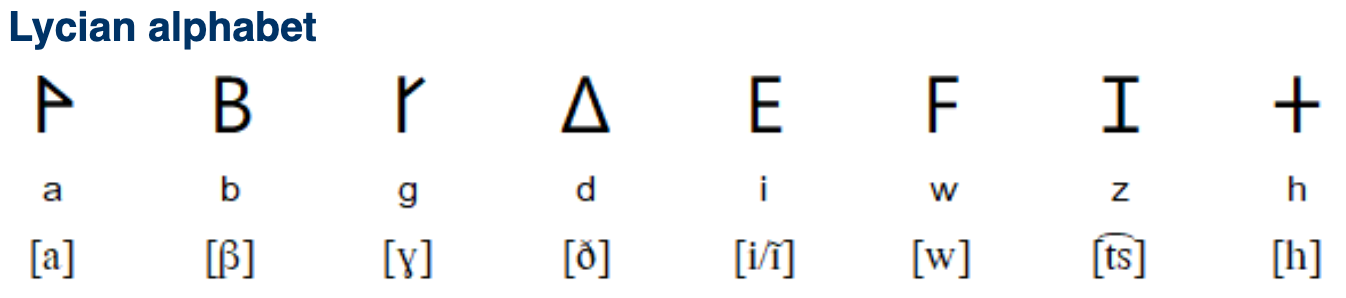}}\par}\\

\midrule

% \textbf{Example 2}

% \textbf{Question}: Which bird(s) in the Wingspan bird cards from the Asia expansion have the word "Great" (but not "Greater") in their name and can reside exclusively in wetlands?

% \textbf{Deep Research Response}: Great Cormorant (Phalacrocorax carbo) – A large waterbird found across Eurasia. In Wingspan Asia, the Great Cormorant can only be played in Wetlands (it has the Wetland habitat icon exclusively on its card) [\href{https://wingspan.rulepop.com/\#:~:text=,Great\%20Indian\%20Bustard}{WINGSPAN.RULEPOP.COM}, \href{https://www.audubon.org/field-guide/bird/great-cormorant\#:~:text=Great\%20Cormorant}{AUDUBON.ORG}]. This satisfies both criteria: its name contains ``Great'' (not ``Greater''), and its sole habitat in the game is Wetlands.

% \textbf{Explanation}: The answer cites Audubon and Wingspan Rulepop, both of which are either irrelevant to the question or fail to support the answer. Based on the step content, the agent most likely retrieved the answer from \url{wingsplain.com} but this source is not cited. \\
% \midrule 

\textbf{Example 3}: \multimodal

\textbf{Question}: When the background of the Patatap website is pink, a circle appears when you press the "C" key on your keyboard. What basic color is the circle?

\textbf{Computer Use Response}: 
I'm unable to complete this task at the moment due to technical limitations. The website appears to be loading but not responding to keyboard inputs as expected.
% Unfortunately, I'm unable to complete this task at the moment due to technical limitations. The website appears to be loading but not responding to keyboard inputs as expected.

\textbf{Explanation}: The agent has limited capabilities of multimodal interactions.\\

\bottomrule

\end{tabular}
}
\caption{Examples of agent errors, corresponding to the discussion in \sectionref{sec:discussion}.}
\label{tab:agent_examples}
\end{table}

% Reasons for Incorrect or Abandoned Answers
% x Incorrect source/tool selection (12, 37, 64)
% x Missed details in the question or answer (16, 21, 23, 28, 71, 101; 78)
% x Lack of topic understanding (18, 32, 68, 104)
% x Obvious oversight (87)

% Reasons for Giving Up
% x Task too complex (45)
% x Lack of topic understanding (49)
% x Conventional way of thinking (91)

Despite advancements in web agents, \name~reveals several critical challenges limiting their effectiveness. These issues span multiple dimensions, including  transparency, source credibility, interaction capabilities, and strategic planning. We discuss each of them below with concrete examples in \tableref{tab:agent_examples} 
% in \appendixref{appendix:agent_errors} 
and agent-specific behavior in \appendixref{appendix:agent_specific_behavior}.

\textbf{Agent developers should enhance trajectories interpretability.} All tested agents provided some access to their action trajectories; however, we notice significant variance in the level of detail present in these trajectories. We recorded the number of steps each agent took per each question and find stark contrasts:\footnote{\label{fn:steps}For \textbf{Grok3 DeepSearch}, we use the number of sources shown in sessions as steps. For \textbf{OpenAI Deep Research}, we extract activity trajectories from saved HTML files of sessions and count steps. For \textbf{Anthropic Computer Use}, we count actions printed on the screen starting with ``\texttt{Tool Use}'' as steps. For \textbf{Convergence AI Proxy}, sessions show the number of steps. Finally, for \textbf{OpenAI Operator}, we copied the steps shown in the session and count steps.} Grok3 DeepSearch provides highly granular reports, averaging 69.8 steps per question; Proxy offers only brief summaries (6.2 steps); Deep Research offers only top-level URLs (see Example 1 in \tableref{tab:agent_examples}).\footnote{We note that the use of brief summaries or obscured URLs may be an intentional anti-distillation strategy used by agent providers to prevent model distillation \citep{savani2025antidistillationsampling}. While this is a plausible explanation, the poor performance of the agents undermines the purpose of anti-distillation sampling which is to be anti-distillation while preserving the teacher models' capabilities.} Such behavior is not desirable---excessive detail obscures key decision-making steps, while overly concise reports and vague URLs reduce transparency. It is harder to evaluate and identify failure points in such situations. As such, we advocate for the release of clear and structured search and reasoning trajectories, which can increase users' trust in agent outputs. 

% \ys{bold, problem description, suggestion}

\textbf{Agents should be evaluated on source credibility.} While most agent responses are grounded in (and attributed to) specific sources, these sources are not always reliable. As shown in \figureref{fig:groundedness} (\appendixref{appendix:correct_answer_source_prop}), OpenAI Deep Research grounds 38.5\% of its correct answers in a unreliable source or no source. In Example 2 (\tableref{tab:agent_examples}), despite successfully locating the source specified in the question, Operator disregards it in favor of an alternative. We hope that future work investigates source credibility more thoroughly, as focusing on correctness only obscures this issue.

\textbf{Agents should enhance and embrace multimodal interactions.} The low accuracy on \name~suggests that the agents either actively avoid interactions or have limited capability with them. In Example 2 (\tableref{tab:agent_examples}), although Operator located the correct source, it avoided navigating through the scanned book. Computer Use, in Example 3, failed to interact with a game. Besides their limited interaction capabilities, agents also faced frequent access denial to content (e.g., videos or reddit posts). While ChatGPT Agent shows significant improvement and approaches human-level performance on \textonly~questions, substantial room for improvement remains, particularly on \multimodal~questions. We recommend improving agent interaction skills (e.g., effective use of the mouse, keyboard, and their combination) and designing strategies to handle restricted content access.

\textbf{Agents should execute tasks with better planning and strategy.} Analysis of agent trajectories shows frequent repetition of unsuccessful actions, such as revisiting webpages where they previously failed to find an answer. The agents also often navigate to irrelevant pages (Example 1 in \tableref{tab:agent_examples}), demonstrating inefficient search behavior. This lack of a clear and focused execution plan leads to an accumulation of irrelevant information, ultimately hindering effective decision-making and retrieval \citep{liu-etal-2024-lost, yang2024sweagent}. We speculate that the development of more structured planning mechanisms could optimize search efficiency, minimize redundant actions, and improve decision-making.

\section{Related work}\label{sec:related_work}

Our work on \name\ contributes to the growing body of evaluating LLM-powered agents. It specifically relates to:

\textbf{Low-level skills:} WebSuite \citep{websuite} and WebGames \citep{webgames} assess fundamental web operations. They identify failure points in complex tasks and offer granular insights into agents' proficiency, albeit with a focus on basic web UI operations.

\textbf{Web agent evaluation:} Web agent evaluation benchmarks broadly fall into text-only and multimodal approaches. The former relies on text-based information: for example, HTML as in Mind2Web \citep{mindweb, webwalker} or various types of text as in WebGPT \citep{webgpt} and WebVoyager \citep{webvoyager}, among others \citep{weblinx, yang2025agentoccam, theagentcompany}. Meanwhile, the latter requires the ability to process multimodal information formats as in VisualWebArena \citep{visualwebarena}, \textsc{TurkingBench} \citep{turkingbench}, and WebArena \citep{webarena}. Closest to our contribution is AssistantBench~\citep{yoran-etal-2024-assistantbench}, which focuses on realistic and time-consuming tasks conducted on the real web. However, while AssistantBench intentionally limits multimodal interactions, such as video understanding, \name~emphasizes diverse multimodal capabilities.

\textbf{Non-web agent evaluation:} ScienceAgentBench \citep{scienceagentbench} evaluates AI agents on scientific discovery, while SWE-Bench \citep{jimenez2024swebench} focuses on software engineering skills. On the other hand, \textsc{OSWorld} \citep{osworld} evaluate AI agents as a generalist for open-ended tasks in real computer environments, similar in spirit to our work with \name.
Agent evaluation is an active field with other diverse focus areas. For example, ST-WebAgentBench \citep{stwebagentbench} examines web agent safety, CowPilot \citep{cowpilot} explores human-agent interactions, and \citet{wei2025browsecompsimplechallengingbenchmark} measures the ability of AI agents to locate hard-to-find information.
% text only: webgpt, weblinx, AgentOccam, TheAgentCompany, WebWalker, Mind2Web (html), WebVoyager

% \textbf{Non-web agent}: ScienceAgentBench, SWE-Bench, Triad, OSWorld

% \textbf{Others} ST-WebAgentBench--safety; CRAB, cross environment tasks, CowPilot

% \url{https://arxiv.org/pdf/2401.13649}, \citet{koh-etal-2024-visualwebarena}

% \url{https://arxiv.org/pdf/2410.13825}, \citet{yang2025agentoccam}

% \url{https://openreview.net/pdf?id=kiYqbO3wqw}, \citet{deng2023mindweb}

% \url{https://openreview.net/pdf?id=VTF8yNQM66}, \citet{jimenez2024swebench}
%
% \url{https://aclanthology.org/2024.emnlp-main.101.pdf}, \citet{zong-etal-2024-triad}

% \url{https://openreview.net/forum?id=6z4YKr0GK6}, \citet{chen2025scienceagentbench}

% \url{https://arxiv.org/abs/2408.04168}, \citet{zeng2024perceivereflectplandesigning}

% \url{https://aclanthology.org/2024.emnlp-main.505/}, \citet{yoran-etal-2024-assistantbench} They proposed a model that is able to plan and replan its steps. This is what our evaluated agents lack

% \url{https://arxiv.org/abs/2402.05930}, \citet{lu2024weblinx}
\section{Conclusion}
\label{sec:conclusion}

We introduce \name, a dataset designed to evaluate the ability of a web agent to identify factual information from the real web through multimodal interactions. Through careful dataset creation and curation, we identify and mitigate key challenges in evaluating web agents, including web contamination, agent workarounds, interaction diversity, and slow evaluation. We find that agents lag significantly behind human performance, particularly on multimodal interactions that humans find trivial to perform. Finally, we highlight impactful directions for future agent development, including enhancing trajectory transparency, source credibility, multimodal capabilities, and planning.

% \begin{enumerate}[label=(\arabic*), itemsep=-0.05em, leftmargin=*]

%     \item \textbf{Enhance transparency}: Agents should make their search and reasoning trajectories more interpretable, benefiting both users and evaluation;

%     \item \textbf{Prioritize source credibility}: Strengthening source reliability is essential for improving answer trustworthiness;

%     \item \textbf{Develop structured execution strategies}: Agents should plan their approach before acting to improve efficiency and accuracy;

%     \item \textbf{Expand evaluation metrics}: Beyond accuracy, benchmarks should assess factuality, reliability, and interaction capabilities;

%     \item \textbf{Evolving Benchmarks}: Researchers should continuously update benchmarks to prevent exploitation by agents with advanced web-searching capabilities.

% \end{enumerate}
\section*{Acknowledgments}

We sincerely thank Meiyu Li and Ali Nirheche for volunteering in the Chinese and Arabic human study. We thank Ben Glickenhaus for helping with the \name~website. We extend gratitude to Kalpesh Krishna, who shared question design ideas with us at the beginning of the project. We thank the Upwork annotators for their hard work, and the members from the UMass NLP lab and UMD CLIP lab for their feedback. This project was partially supported by awards IIS2046248, IIS-2312949, and IIS-2202506 from the National Science Foundation (NSF).

\bibliography{colm2025_conference, custom}

\begin{thebibliography}{41}
\providecommand{\natexlab}[1]{#1}
\providecommand{\url}[1]{\texttt{#1}}
\expandafter\ifx\csname urlstyle\endcsname\relax
  \providecommand{\doi}[1]{doi: #1}\else
  \providecommand{\doi}{doi: \begingroup \urlstyle{rm}\Url}\fi

\bibitem[{Anthropic}(2024)]{anthropic2024}
{Anthropic}.
\newblock {Introducing computer use, a new Claude 3.5 Sonnet, and Claude 3.5 Haiku}, 2024.
\newblock URL \url{https://www.anthropic.com/news/3-5-models-and-computer-use}.
\newblock Accessed: March 4, 2025.

\bibitem[Chen et~al.(2025)Chen, Chen, Ning, Zhang, Wang, Yu, Li, Liao, Wei, Lu, Dey, Xue, Baker, Burns, Adu-Ampratwum, Huang, Ning, Gao, Su, and Sun]{scienceagentbench}
Ziru Chen, Shijie Chen, Yuting Ning, Qianheng Zhang, Boshi Wang, Botao Yu, Yifei Li, Zeyi Liao, Chen Wei, Zitong Lu, Vishal Dey, Mingyi Xue, Frazier~N. Baker, Benjamin Burns, Daniel Adu-Ampratwum, Xuhui Huang, Xia Ning, Song Gao, Yu~Su, and Huan Sun.
\newblock {ScienceAgentBench}: Toward rigorous assessment of language agents for data-driven scientific discovery.
\newblock In \emph{The Thirteenth International Conference on Learning Representations}, 2025.
\newblock URL \url{https://openreview.net/forum?id=6z4YKr0GK6}.

\bibitem[{Convergence AI}(2025)]{convergenceAI}
{Convergence AI}.
\newblock {Proxy: Your AI assistant for your daily tasks}, 2025.
\newblock URL \url{https://convergence.ai/}.
\newblock Accessed: March 4, 2025.

\bibitem[DeepSeek-AI et~al.(2025)DeepSeek-AI, Liu, Feng, Xue, Wang, Wu, Lu, Zhao, Deng, Zhang, Ruan, Dai, Guo, Yang, Chen, Ji, Li, Lin, Dai, Luo, Hao, Chen, Li, Zhang, Bao, Xu, Wang, Zhang, Ding, Xin, Gao, Li, Qu, Cai, Liang, Guo, Ni, Li, Wang, Chen, Chen, Yuan, Qiu, Li, Song, Dong, Hu, Gao, Guan, Huang, Yu, Wang, Zhang, Xu, Xia, Zhao, Wang, Zhang, Li, Wang, Zhang, Zhang, Tang, Li, Tian, Huang, Wang, Zhang, Wang, Zhu, Chen, Du, Chen, Jin, Ge, Zhang, Pan, Wang, Xu, Zhang, Chen, Li, Lu, Zhou, Chen, Wu, Ye, Ye, Ma, Wang, Zhou, Yu, Zhou, Pan, Wang, Yun, Pei, Sun, Xiao, Zeng, Zhao, An, Liu, Liang, Gao, Yu, Zhang, Li, Jin, Wang, Bi, Liu, Wang, Shen, Chen, Zhang, Chen, Nie, Sun, Wang, Cheng, Liu, Xie, Liu, Yu, Song, Shan, Zhou, Yang, Li, Su, Lin, Li, Wang, Wei, Zhu, Zhang, Xu, Xu, Huang, Li, Zhao, Sun, Li, Wang, Yu, Zheng, Zhang, Shi, Xiong, He, Tang, Piao, Wang, Tan, Ma, Liu, Guo, Wu, Ou, Zhu, Wang, Gong, Zou, He, Zha, Xiong, Ma, Yan, Luo, You, Liu, Zhou, Wu, Ren, Ren, Sha, Fu, Xu, Huang, Zhang, Xie, Zhang, Hao,
  Gou, Ma, Yan, Shao, Xu, Wu, Zhang, Li, Gu, Zhu, Liu, Li, Xie, Song, Gao, and Pan]{deepseekai2025deepseekv3technicalreport}
DeepSeek-AI, Aixin Liu, Bei Feng, Bing Xue, Bingxuan Wang, Bochao Wu, Chengda Lu, Chenggang Zhao, Chengqi Deng, Chenyu Zhang, Chong Ruan, Damai Dai, Daya Guo, Dejian Yang, Deli Chen, Dongjie Ji, Erhang Li, Fangyun Lin, Fucong Dai, Fuli Luo, Guangbo Hao, Guanting Chen, Guowei Li, H.~Zhang, Han Bao, Hanwei Xu, Haocheng Wang, Haowei Zhang, Honghui Ding, Huajian Xin, Huazuo Gao, Hui Li, Hui Qu, J.~L. Cai, Jian Liang, Jianzhong Guo, Jiaqi Ni, Jiashi Li, Jiawei Wang, Jin Chen, Jingchang Chen, Jingyang Yuan, Junjie Qiu, Junlong Li, Junxiao Song, Kai Dong, Kai Hu, Kaige Gao, Kang Guan, Kexin Huang, Kuai Yu, Lean Wang, Lecong Zhang, Lei Xu, Leyi Xia, Liang Zhao, Litong Wang, Liyue Zhang, Meng Li, Miaojun Wang, Mingchuan Zhang, Minghua Zhang, Minghui Tang, Mingming Li, Ning Tian, Panpan Huang, Peiyi Wang, Peng Zhang, Qiancheng Wang, Qihao Zhu, Qinyu Chen, Qiushi Du, R.~J. Chen, R.~L. Jin, Ruiqi Ge, Ruisong Zhang, Ruizhe Pan, Runji Wang, Runxin Xu, Ruoyu Zhang, Ruyi Chen, S.~S. Li, Shanghao Lu, Shangyan Zhou, Shanhuang
  Chen, Shaoqing Wu, Shengfeng Ye, Shengfeng Ye, Shirong Ma, Shiyu Wang, Shuang Zhou, Shuiping Yu, Shunfeng Zhou, Shuting Pan, T.~Wang, Tao Yun, Tian Pei, Tianyu Sun, W.~L. Xiao, Wangding Zeng, Wanjia Zhao, Wei An, Wen Liu, Wenfeng Liang, Wenjun Gao, Wenqin Yu, Wentao Zhang, X.~Q. Li, Xiangyue Jin, Xianzu Wang, Xiao Bi, Xiaodong Liu, Xiaohan Wang, Xiaojin Shen, Xiaokang Chen, Xiaokang Zhang, Xiaosha Chen, Xiaotao Nie, Xiaowen Sun, Xiaoxiang Wang, Xin Cheng, Xin Liu, Xin Xie, Xingchao Liu, Xingkai Yu, Xinnan Song, Xinxia Shan, Xinyi Zhou, Xinyu Yang, Xinyuan Li, Xuecheng Su, Xuheng Lin, Y.~K. Li, Y.~Q. Wang, Y.~X. Wei, Y.~X. Zhu, Yang Zhang, Yanhong Xu, Yanhong Xu, Yanping Huang, Yao Li, Yao Zhao, Yaofeng Sun, Yaohui Li, Yaohui Wang, Yi~Yu, Yi~Zheng, Yichao Zhang, Yifan Shi, Yiliang Xiong, Ying He, Ying Tang, Yishi Piao, Yisong Wang, Yixuan Tan, Yiyang Ma, Yiyuan Liu, Yongqiang Guo, Yu~Wu, Yuan Ou, Yuchen Zhu, Yuduan Wang, Yue Gong, Yuheng Zou, Yujia He, Yukun Zha, Yunfan Xiong, Yunxian Ma, Yuting Yan, Yuxiang
  Luo, Yuxiang You, Yuxuan Liu, Yuyang Zhou, Z.~F. Wu, Z.~Z. Ren, Zehui Ren, Zhangli Sha, Zhe Fu, Zhean Xu, Zhen Huang, Zhen Zhang, Zhenda Xie, Zhengyan Zhang, Zhewen Hao, Zhibin Gou, Zhicheng Ma, Zhigang Yan, Zhihong Shao, Zhipeng Xu, Zhiyu Wu, Zhongyu Zhang, Zhuoshu Li, Zihui Gu, Zijia Zhu, Zijun Liu, Zilin Li, Ziwei Xie, Ziyang Song, Ziyi Gao, and Zizheng Pan.
\newblock Deepseek-v3 technical report, 2025.
\newblock URL \url{https://arxiv.org/abs/2412.19437}.

\bibitem[Deng et~al.(2023)Deng, Gu, Zheng, Chen, Stevens, Wang, Sun, and Su]{mindweb}
Xiang Deng, Yu~Gu, Boyuan Zheng, Shijie Chen, Samuel Stevens, Boshi Wang, Huan Sun, and Yu~Su.
\newblock {Mind2Web}: Towards a generalist agent for the web.
\newblock In \emph{Thirty-seventh Conference on Neural Information Processing Systems Datasets and Benchmarks Track}, 2023.
\newblock URL \url{https://openreview.net/forum?id=kiYqbO3wqw}.

\bibitem[{Google Gemini}(2024)]{google2024geminideepresearch}
{Google Gemini}.
\newblock Gemini deep research.
\newblock \url{https://gemini.google/overview/deep-research/}, December 2024.
\newblock Accessed July 2025.

\bibitem[He et~al.(2024)He, Yao, Ma, Yu, Dai, Zhang, Lan, and Yu]{webvoyager}
Hongliang He, Wenlin Yao, Kaixin Ma, Wenhao Yu, Yong Dai, Hongming Zhang, Zhenzhong Lan, and Dong Yu.
\newblock {W}eb{V}oyager: Building an end-to-end web agent with large multimodal models.
\newblock In Lun-Wei Ku, Andre Martins, and Vivek Srikumar (eds.), \emph{Proceedings of the 62nd Annual Meeting of the Association for Computational Linguistics (Volume 1: Long Papers)}, pp.\  6864--6890, Bangkok, Thailand, August 2024. Association for Computational Linguistics.
\newblock \doi{10.18653/v1/2024.acl-long.371}.
\newblock URL \url{https://aclanthology.org/2024.acl-long.371/}.

\bibitem[Huq et~al.(2025)Huq, Wang, Xu, Ou, Zhou, Bigham, and Neubig]{cowpilot}
Faria Huq, Zora~Zhiruo Wang, Frank~F. Xu, Tianyue Ou, Shuyan Zhou, Jeffrey~P. Bigham, and Graham Neubig.
\newblock {CowPilot}: A framework for autonomous and human-agent collaborative web navigation, 2025.
\newblock URL \url{https://arxiv.org/abs/2501.16609}.

\bibitem[Jimenez et~al.(2024)Jimenez, Yang, Wettig, Yao, Pei, Press, and Narasimhan]{jimenez2024swebench}
Carlos~E Jimenez, John Yang, Alexander Wettig, Shunyu Yao, Kexin Pei, Ofir Press, and Karthik~R Narasimhan.
\newblock {SWE}-bench: Can language models resolve real-world github issues?
\newblock In \emph{The Twelfth International Conference on Learning Representations}, 2024.
\newblock URL \url{https://openreview.net/forum?id=VTF8yNQM66}.

\bibitem[Karpinska et~al.(2024)Karpinska, Thai, Lo, Goyal, and Iyyer]{karpinska-etal-2024-one}
Marzena Karpinska, Katherine Thai, Kyle Lo, Tanya Goyal, and Mohit Iyyer.
\newblock One thousand and one pairs: A {\textquotedblleft}novel{\textquotedblright} challenge for long-context language models.
\newblock In Yaser Al-Onaizan, Mohit Bansal, and Yun-Nung Chen (eds.), \emph{Proceedings of the 2024 Conference on Empirical Methods in Natural Language Processing}, pp.\  17048--17085, Miami, Florida, USA, November 2024. Association for Computational Linguistics.
\newblock \doi{10.18653/v1/2024.emnlp-main.948}.
\newblock URL \url{https://aclanthology.org/2024.emnlp-main.948/}.

\bibitem[Koh et~al.(2024)Koh, Lo, Jang, Duvvur, Lim, Huang, Neubig, Zhou, Salakhutdinov, and Fried]{visualwebarena}
Jing~Yu Koh, Robert Lo, Lawrence Jang, Vikram Duvvur, Ming Lim, Po-Yu Huang, Graham Neubig, Shuyan Zhou, Russ Salakhutdinov, and Daniel Fried.
\newblock {V}isual{W}eb{A}rena: Evaluating multimodal agents on realistic visual web tasks.
\newblock In Lun-Wei Ku, Andre Martins, and Vivek Srikumar (eds.), \emph{Proceedings of the 62nd Annual Meeting of the Association for Computational Linguistics (Volume 1: Long Papers)}, pp.\  881--905, Bangkok, Thailand, August 2024. Association for Computational Linguistics.
\newblock \doi{10.18653/v1/2024.acl-long.50}.
\newblock URL \url{https://aclanthology.org/2024.acl-long.50/}.

\bibitem[Levy et~al.(2025)Levy, wiesel, Marreed, Oved, Yaeli, and Shlomov]{stwebagentbench}
Ido Levy, Ben wiesel, Sami Marreed, Alon Oved, Avi Yaeli, and Segev Shlomov.
\newblock {ST}-{WebAgentBench}: A benchmark for evaluating safety and trustworthiness in web agents, 2025.
\newblock URL \url{https://openreview.net/forum?id=IIzehISTBe}.

\bibitem[Li \& Waldo(2024)Li and Waldo]{websuite}
Eric Li and Jim Waldo.
\newblock {WebSuite}: Systematically evaluating why web agents fail, 2024.
\newblock URL \url{https://arxiv.org/abs/2406.01623}.

\bibitem[Liu et~al.(2024)Liu, Lin, Hewitt, Paranjape, Bevilacqua, Petroni, and Liang]{liu-etal-2024-lost}
Nelson~F. Liu, Kevin Lin, John Hewitt, Ashwin Paranjape, Michele Bevilacqua, Fabio Petroni, and Percy Liang.
\newblock Lost in the middle: How language models use long contexts.
\newblock \emph{Transactions of the Association for Computational Linguistics}, 12:\penalty0 157--173, 2024.
\newblock \doi{10.1162/tacl_a_00638}.
\newblock URL \url{https://aclanthology.org/2024.tacl-1.9/}.

\bibitem[Lù et~al.(2024)Lù, Kasner, and Reddy]{weblinx}
Xing~Han Lù, Zdeněk Kasner, and Siva Reddy.
\newblock Weblinx: Real-world website navigation with multi-turn dialogue, 2024.
\newblock URL \url{https://arxiv.org/abs/2402.05930}.

\bibitem[{MAA}(2024)]{AIME2024}
{MAA}.
\newblock American invitational mathematics examination - aime, February 2024.
\newblock URL \url{https://maa.org/math-competitions/american-invitational-mathematics-examination-aime, https://huggingface.co/datasets/Maxwell-Jia/AIME_2024}.

\bibitem[Nakano et~al.(2022)Nakano, Hilton, Balaji, Wu, Ouyang, Kim, Hesse, Jain, Kosaraju, Saunders, Jiang, Cobbe, Eloundou, Krueger, Button, Knight, Chess, and Schulman]{webgpt}
Reiichiro Nakano, Jacob Hilton, Suchir Balaji, Jeff Wu, Long Ouyang, Christina Kim, Christopher Hesse, Shantanu Jain, Vineet Kosaraju, William Saunders, Xu~Jiang, Karl Cobbe, Tyna Eloundou, Gretchen Krueger, Kevin Button, Matthew Knight, Benjamin Chess, and John Schulman.
\newblock {WebGPT}: Browser-assisted question-answering with human feedback, 2022.
\newblock URL \url{https://arxiv.org/abs/2112.09332}.

\bibitem[OpenAI(2024)]{OpenAI_ComputerUsingAgent_2024}
OpenAI.
\newblock Computer-using agent, 2024.
\newblock URL \url{https://openai.com/index/computer-using-agent/}.
\newblock Accessed: 2025-03-06.

\bibitem[OpenAI(2025{\natexlab{a}})]{openai2024chatgptagent}
OpenAI.
\newblock Introducing chatgpt agent: bridging research and action, 2025{\natexlab{a}}.
\newblock URL \url{https://openai.com/index/introducing-chatgpt-agent//}.
\newblock Accessed: 2025-07-21.

\bibitem[OpenAI(2025{\natexlab{b}})]{openai2025deepresearch}
OpenAI.
\newblock Deep research system card, February 2025{\natexlab{b}}.
\newblock URL \url{https://cdn.openai.com/deep-research-system-card.pdf}.
\newblock Accessed: 2025-02-27.

\bibitem[{OpenAI}(2025)]{openai2025o3o4mini}
{OpenAI}.
\newblock Introducing openai o3 and o4‑mini.
\newblock \url{https://openai.com/index/introducing-o3-and-o4-mini/}, April 2025.
\newblock Accessed July 2025.

\bibitem[OpenAI(2025)]{openai2025operator}
OpenAI.
\newblock Operator system card, January 2025.
\newblock URL \url{https://cdn.openai.com/operator_system_card.pdf}.
\newblock Accessed: 2025-03-01.

\bibitem[{Perplexity AI}(2025)]{perplexitySonarPro}
{Perplexity AI}.
\newblock Build with the best ai answer engine.
\newblock \url{https://sonar.perplexity.ai/}, April 2025.
\newblock Accessed July 2025.

\bibitem[Petrov et~al.(2025)Petrov, Dekoninck, Baltadzhiev, Drencheva, Minchev, Balunović, Jovanović, and Vechev]{petrov2025proofbluffevaluatingllms}
Ivo Petrov, Jasper Dekoninck, Lyuben Baltadzhiev, Maria Drencheva, Kristian Minchev, Mislav Balunović, Nikola Jovanović, and Martin Vechev.
\newblock Proof or bluff? evaluating llms on 2025 usa math olympiad, 2025.
\newblock URL \url{https://arxiv.org/abs/2503.21934}.

\bibitem[Quan et~al.(2025)Quan, Yang, Yu, Zheng, Liu, Yang, Ren, Gao, Miao, Feng, et~al.]{quan2025codeelo}
Shanghaoran Quan, Jiaxi Yang, Bowen Yu, Bo~Zheng, Dayiheng Liu, An~Yang, Xuancheng Ren, Bofei Gao, Yibo Miao, Yunlong Feng, et~al.
\newblock Codeelo: Benchmarking competition-level code generation of llms with human-comparable elo ratings.
\newblock \emph{arXiv preprint arXiv:2501.01257}, 2025.

\bibitem[Rein et~al.(2024)Rein, Hou, Stickland, Petty, Pang, Dirani, Michael, and Bowman]{rein2024gpqa}
David Rein, Betty~Li Hou, Asa~Cooper Stickland, Jackson Petty, Richard~Yuanzhe Pang, Julien Dirani, Julian Michael, and Samuel~R. Bowman.
\newblock {GPQA}: A graduate-level google-proof q\&a benchmark.
\newblock In \emph{First Conference on Language Modeling}, 2024.
\newblock URL \url{https://openreview.net/forum?id=Ti67584b98}.

\bibitem[Sainz et~al.(2023)Sainz, Campos, Garc{\'i}a-Ferrero, Etxaniz, de~Lacalle, and Agirre]{sainz-etal-2023-nlp}
Oscar Sainz, Jon Campos, Iker Garc{\'i}a-Ferrero, Julen Etxaniz, Oier~Lopez de~Lacalle, and Eneko Agirre.
\newblock {NLP} evaluation in trouble: On the need to measure {LLM} data contamination for each benchmark.
\newblock In Houda Bouamor, Juan Pino, and Kalika Bali (eds.), \emph{Findings of the Association for Computational Linguistics: EMNLP 2023}, Singapore, December 2023.

\bibitem[Savani et~al.(2025)Savani, Trockman, Feng, Schwarzschild, Robey, Finzi, and Kolter]{savani2025antidistillationsampling}
Yash Savani, Asher Trockman, Zhili Feng, Avi Schwarzschild, Alexander Robey, Marc Finzi, and J.~Zico Kolter.
\newblock Antidistillation sampling, 2025.
\newblock URL \url{https://arxiv.org/abs/2504.13146}.

\bibitem[Thomas et~al.(2025)Thomas, Chan, Kang, Wu, Christianos, Greenlee, Toulis, and Purtorab]{webgames}
George Thomas, Alex~J. Chan, Jikun Kang, Wenqi Wu, Filippos Christianos, Fraser Greenlee, Andy Toulis, and Marvin Purtorab.
\newblock {WebGames}: Challenging general-purpose web-browsing ai agents, 2025.
\newblock URL \url{https://arxiv.org/abs/2502.18356}.

\bibitem[Vu et~al.(2024)Vu, Iyyer, Wang, Constant, Wei, Wei, Tar, Sung, Zhou, Le, and Luong]{vu-etal-2024-freshllms}
Tu~Vu, Mohit Iyyer, Xuezhi Wang, Noah Constant, Jerry Wei, Jason Wei, Chris Tar, Yun-Hsuan Sung, Denny Zhou, Quoc Le, and Thang Luong.
\newblock {F}resh{LLM}s: Refreshing large language models with search engine augmentation.
\newblock In Lun-Wei Ku, Andre Martins, and Vivek Srikumar (eds.), \emph{Findings of the Association for Computational Linguistics: ACL 2024}, pp.\  13697--13720, Bangkok, Thailand, August 2024. Association for Computational Linguistics.
\newblock \doi{10.18653/v1/2024.findings-acl.813}.
\newblock URL \url{https://aclanthology.org/2024.findings-acl.813/}.

\bibitem[Wei et~al.(2025)Wei, Sun, Papay, McKinney, Han, Fulford, Chung, Passos, Fedus, and Glaese]{wei2025browsecompsimplechallengingbenchmark}
Jason Wei, Zhiqing Sun, Spencer Papay, Scott McKinney, Jeffrey Han, Isa Fulford, Hyung~Won Chung, Alex~Tachard Passos, William Fedus, and Amelia Glaese.
\newblock Browsecomp: A simple yet challenging benchmark for browsing agents, 2025.
\newblock URL \url{https://arxiv.org/abs/2504.12516}.

\bibitem[Wu et~al.(2025)Wu, Yin, Jiang, Wang, Xi, Fang, Zhang, He, Zhou, Xie, and Huang]{webwalker}
Jialong Wu, Wenbiao Yin, Yong Jiang, Zhenglin Wang, Zekun Xi, Runnan Fang, Linhai Zhang, Yulan He, Deyu Zhou, Pengjun Xie, and Fei Huang.
\newblock {WebWalker}: Benchmarking llms in web traversal, 2025.
\newblock URL \url{https://arxiv.org/abs/2501.07572}.

\bibitem[Xie et~al.(2024)Xie, Zhang, Chen, Li, Zhao, Cao, Hua, Cheng, Shin, Lei, Liu, Xu, Zhou, Savarese, Xiong, Zhong, and Yu]{osworld}
Tianbao Xie, Danyang Zhang, Jixuan Chen, Xiaochuan Li, Siheng Zhao, Ruisheng Cao, Toh~Jing Hua, Zhoujun Cheng, Dongchan Shin, Fangyu Lei, Yitao Liu, Yiheng Xu, Shuyan Zhou, Silvio Savarese, Caiming Xiong, Victor Zhong, and Tao Yu.
\newblock {OSWorld}: Benchmarking multimodal agents for open-ended tasks in real computer environments, 2024.
\newblock URL \url{https://arxiv.org/abs/2404.07972}.

\bibitem[Xu et~al.(2024)Xu, Song, Li, Tang, Jain, Bao, Wang, Zhou, Guo, Cao, Yang, Lu, Martin, Su, Maben, Mehta, Chi, Jang, Xie, Zhou, and Neubig]{theagentcompany}
Frank~F. Xu, Yufan Song, Boxuan Li, Yuxuan Tang, Kritanjali Jain, Mengxue Bao, Zora~Z. Wang, Xuhui Zhou, Zhitong Guo, Murong Cao, Mingyang Yang, Hao~Yang Lu, Amaad Martin, Zhe Su, Leander Maben, Raj Mehta, Wayne Chi, Lawrence Jang, Yiqing Xie, Shuyan Zhou, and Graham Neubig.
\newblock {TheAgentCompany}: Benchmarking llm agents on consequential real world tasks, 2024.
\newblock URL \url{https://arxiv.org/abs/2412.14161}.

\bibitem[Xu et~al.(2025)Xu, Kordi, Nayak, Asija, Wang, Sanders, Byerly, Zhang, Durme, and Khashabi]{turkingbench}
Kevin Xu, Yeganeh Kordi, Tanay Nayak, Adi Asija, Yizhong Wang, Kate Sanders, Adam Byerly, Jingyu Zhang, Benjamin~Van Durme, and Daniel Khashabi.
\newblock {Tur[k]ingBench}: A challenge benchmark for web agents, 2025.
\newblock URL \url{https://arxiv.org/abs/2403.11905}.

\bibitem[Yang et~al.(2024)Yang, Jimenez, Wettig, Lieret, Yao, Narasimhan, and Press]{yang2024sweagent}
John Yang, Carlos~E Jimenez, Alexander Wettig, Kilian Lieret, Shunyu Yao, Karthik~R Narasimhan, and Ofir Press.
\newblock {SWE}-agent: Agent-computer interfaces enable automated software engineering.
\newblock In \emph{The Thirty-eighth Annual Conference on Neural Information Processing Systems}, 2024.
\newblock URL \url{https://openreview.net/forum?id=mXpq6ut8J3}.

\bibitem[Yang et~al.(2025)Yang, Liu, Chaudhary, Fakoor, Chaudhari, Karypis, and Rangwala]{yang2025agentoccam}
Ke~Yang, Yao Liu, Sapana Chaudhary, Rasool Fakoor, Pratik Chaudhari, George Karypis, and Huzefa Rangwala.
\newblock {AgentOccam}: A simple yet strong baseline for {LLM}-based web agents.
\newblock In \emph{The Thirteenth International Conference on Learning Representations}, 2025.
\newblock URL \url{https://openreview.net/forum?id=oWdzUpOlkX}.

\bibitem[Yao et~al.(2022)Yao, Chen, Yang, and Narasimhan]{webshop}
Shunyu Yao, Howard Chen, John Yang, and Karthik Narasimhan.
\newblock {WebShop}: Towards scalable real-world web interaction with grounded language agents.
\newblock In S.~Koyejo, S.~Mohamed, A.~Agarwal, D.~Belgrave, K.~Cho, and A.~Oh (eds.), \emph{Advances in Neural Information Processing Systems}, volume~35, pp.\  20744--20757. Curran Associates, Inc., 2022.
\newblock URL \url{https://proceedings.neurips.cc/paper_files/paper/2022/file/82ad13ec01f9fe44c01cb91814fd7b8c-Paper-Conference.pdf}.

\bibitem[Ying et~al.(2025)Ying, Collins, Wong, Sucholutsky, Liu, Weller, Shu, Griffiths, and Tenenbaum]{benchmarkinghumanlikeintelligencemachines}
Lance Ying, Katherine~M. Collins, Lionel Wong, Ilia Sucholutsky, Ryan Liu, Adrian Weller, Tianmin Shu, Thomas~L. Griffiths, and Joshua~B. Tenenbaum.
\newblock On benchmarking human-like intelligence in machines, 2025.
\newblock URL \url{https://arxiv.org/abs/2502.20502}.

\bibitem[Yoran et~al.(2024)Yoran, Amouyal, Malaviya, Bogin, Press, and Berant]{yoran-etal-2024-assistantbench}
Ori Yoran, Samuel~Joseph Amouyal, Chaitanya Malaviya, Ben Bogin, Ofir Press, and Jonathan Berant.
\newblock {A}ssistant{B}ench: Can web agents solve realistic and time-consuming tasks?
\newblock In Yaser Al-Onaizan, Mohit Bansal, and Yun-Nung Chen (eds.), \emph{Proceedings of the 2024 Conference on Empirical Methods in Natural Language Processing}, pp.\  8938--8968, Miami, Florida, USA, November 2024. Association for Computational Linguistics.
\newblock \doi{10.18653/v1/2024.emnlp-main.505}.
\newblock URL \url{https://aclanthology.org/2024.emnlp-main.505/}.

\bibitem[Zhou et~al.(2024)Zhou, Xu, Zhu, Zhou, Lo, Sridhar, Cheng, Ou, Bisk, Fried, Alon, and Neubig]{webarena}
Shuyan Zhou, Frank~F. Xu, Hao Zhu, Xuhui Zhou, Robert Lo, Abishek Sridhar, Xianyi Cheng, Tianyue Ou, Yonatan Bisk, Daniel Fried, Uri Alon, and Graham Neubig.
\newblock {WebArena}: A realistic web environment for building autonomous agents.
\newblock In \emph{The Twelfth International Conference on Learning Representations}, 2024.
\newblock URL \url{https://openreview.net/forum?id=oKn9c6ytLx}.

\end{thebibliography}
\bibliographystyle{colm2025_conference}

\appendix

\section{Limitations}

While we ensure high-quality of \name~via rigorous data revision and filtering, we identify the following limitations of the benchmark and hope future work will improve on these aspects to develop more robust and versatile web agents.
First, every question in \name~has a single short answer, whereas in a more realistic setting, some questions may not have an answer at all or may have multiple or even long-form answers. For such questions, agents should provide credible sources for each possible answer and should be evaluated on source quality.
Second, while \name\ includes questions that are multilingual, testing the ability of agents to handle multilingual queries is not the primary goal of our benchmark due to its limited size. We encourage future research to conduct systematic studies on agents' performance across different cultures and languages.
Third, directly comparing agents based on their action trajectories is challenging due to inconsistencies in the level of detail individual agents provide. We encourage future advancements in agent transparency to enable more straightforward and meaningful comparisons.

\section{Data creation workflow}\label{appendix:data_creation_workflow}

\begin{figure}[h!]
\begin{center}
\includegraphics[scale=0.44]{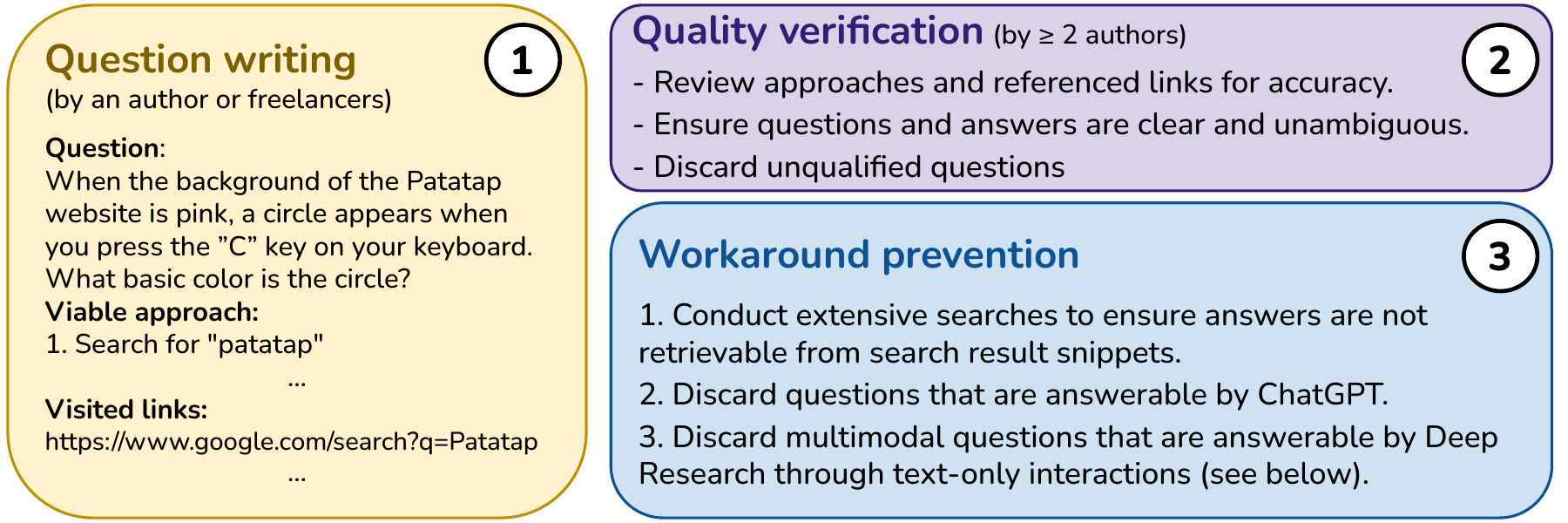}
\end{center}
\caption{Workflow of creating \name. Each question and its viable trajectory and visited links are verified by at least two authors. Each \multimodal~question is verified to be only solvable via multi-modal interactions.}
\label{fig:data_creation_workflow}
\end{figure}

We seek questions that satisfy the following four high-level criteria. The concrete data creation workflow can be found in \figureref{fig:data_creation_workflow}.

\begin{enumerate}[label=(\arabic*), itemsep=-0.05em, leftmargin=*]
    \item \textit{Questions should be short but unambiguous}: questions should provide sufficient yet minimal information to unambiguously lead to correct answers.

    \item \textit{Answers should be trivial to evaluate}: answers must be correct, unique, and concise. Answers cannot be lists or sets, unlike, for example, AssistantBench~\citep{yoran-etal-2024-assistantbench}, and paraphrases of answers cannot be considered correct.

    \item \textit{Answers should be adversarial to Google Search}: answers must not appear in Google Search snippets or top-ranked results when the question or fragments of the question are used as the query. Furthermore, \multimodal\ questions must not be solvable by methods that only operate on text (e.g., Deep Research). 

    \item \textit{Answers must be publicly accessible}: answers must be available on non-paywalled websites, without requiring any account creation or login actions.
\end{enumerate}

\figureref{fig:data_creation_workflow} details the data validation process discussed in \sectionref{sec:dataset}. Each question is verified by at least two authors who review each question carefully against the four criteria, along with its viable trajectory and visited links. We remove questions whose trajectories involve \multimodal\ interaction but were found to have a text-only workaround by non-multimodal agents. 

\section{Human annotator recruitment}\label{appendix:human_recruitment_details}

This section provides details on human annotator recruitment for the human study in \sectionref{sec:human_setup}. We organize \name\ questions into sets, each containing all questions from a specific non-English language and some English questions. We recruit native speaker volunteers for Arabic and Chinese questions, and we hire three annotators via Upwork to handle the Hindi, German, and Finnish questions, respectively. An English-only set is attempted by annotators and authors who did not write or validate them. The hired annotators receive \$2.5 USD per question, with an additional \$1 USD bonus for each correct answer to incentivize accuracy.  

\section{Baseline prompt}\label{appendix:baseline_prompt}

We provide the baseline hyperparameters and prompt for the Google-search-augmented baselines in \tableref{tab:serper_prompt}

\begin{table}[h]
\centering
\fontsize{7}{9}\selectfont
\resizebox{0.9\columnwidth}{!}{%
\begin{tabular}{p{10cm}}
\toprule
\textbf{Model}: \texttt{gpt-4o-2024-11-20} \quad 
\textbf{max\_tokens}: 518 \quad 
\textbf{temperature}: 0 \textcolor{green!80}{\# for both baseline settings} \\
\textbf{Model}: \texttt{DeepSeek R1} \hspace{0.93cm}
\textbf{max\_tokens}: 8000 \hspace{0.07cm}
\textbf{temperature}: 0 \textcolor{green!80}{\# for both baseline settings} \\

\textbf{Prompt}: \textcolor{green!80}{\# for the Google-search-augmented setting}

Use the provided context to answer the question accurately and concisely. Do not use your own knowledge.

- If the context contains a direct answer, provide it in a precise and straightforward manner.

- If the context does not provide a clear answer, reply with 'No answer found'.

- Avoid unnecessary elaboration.

Question: \{question\}

Context: \{context\} \textcolor{green!80}{\# consists of a list of result titles and snippets}

Answer: \\

\bottomrule
\end{tabular}%
}
\caption{Baseline hyperparameters and the prompt in the Google-search-augmented setting.}
\label{tab:serper_prompt}
\end{table}

\section{Human performance results}\label{appendix:human_results_detailed}

\tableref{tab:human_performance} complements the discussion in \sectionref{sec:human_eval_results} by providing detailed results on the number of dead ends encountered, time taken, and perceived difficulty experienced by the human annotators.

% \begin{table}[ht]
% \fontsize{6}{7}\selectfont
% \centering
% \resizebox{0.95\columnwidth}{!}{%
% \begin{tabular}{@{}rp{0.4cm}p{0.35cm}p{0.3cm}p{0.85cm}p{0.2cm}p{0.2cm}p{0.4cm}p{0.2cm}p{0.2cm}p{0.4cm}p{0.2cm}p{0.5cm}p{0.5cm}}
% \toprule
% \multirow{2}{*}{} & \multicolumn{4}{c}{\textbf{Correctness}} & \multicolumn{3}{c}{\textbf{Dead End}} & \multicolumn{3}{c}{\textbf{Time (min:sec)}} & \multicolumn{3}{c}{\textbf{Difficulty label}} \\
% \cmidrule(lr){2-5}
% \cmidrule(lr){6-8}
% \cmidrule(lr){9-11}
% \cmidrule(lr){12-14}
%  & Correct & Wrong & None & Accuracy & Avg. & Min & Max & Avg. & Min & Max  & Easy & Medium & Hard \\
% \midrule
% \textonly  & 49   & 9     & 1   & 83.05\%  & 1.75 & 0   & 12  & 5:10  & 0:30  & 27:24 & 25   & 21     & 13   \\
% \addlinespace[1pt]

% \multimodal & 57   & 6     & 2   & 87.69\%  & 1.22 & 0   & 14  & 4:25  & 0:26  & 24:14 & 36   & 21     & 8    \\
% \addlinespace[1pt]

% All         & 106 & 15    & 3  & 85.48\% & 1.44 & 0   & 14  & 4:42 & 0:26  & 27:24 & 61   & 42     & 21  \\
% \bottomrule
% \end{tabular}
% }
% \caption{Human performance on \name. `None' means that the annotators abandoned a question after 15 minutes. Human achieves 85.5\% accuracy (\textonly~83.1\% and \multimodal~87.7\%). Humans in general were able to find answers smoothly (1.44 dead end on average) and quickly (4 min 42 sec on average).}
% \label{tab:human_performance}
% \end{table}
% % https://docs.google.com/spreadsheets/d/1vrq9D1eXMFKcH7pv9etdYVx6Uoi-xduo0vMw7NsUssI/edit?usp=sharing

\begin{table}[h]
\fontsize{6}{7}\selectfont
\centering
\resizebox{0.95\columnwidth}{!}{%
\begin{tabular}{@{}rp{0.6cm}p{0.35cm}p{0.3cm}p{0.5cm}p{0.2cm}p{0.4cm}p{0.2cm}p{0.2cm}p{0.4cm}p{0.2cm}p{0.5cm}p{0.5cm}}
\toprule
\multirow{2}{*}{} & \multicolumn{4}{c}{\textbf{Correctness}} & \multicolumn{2}{c}{\textbf{Dead End}} & \multicolumn{3}{c}{\textbf{Time (min:sec)}} & \multicolumn{3}{c}{\textbf{Perceived Difficulty}} \\
\cmidrule(lr){2-5}
\cmidrule(lr){6-7}
\cmidrule(lr){8-10}
\cmidrule(lr){11-13}
 & Accuracy & Correct & Wrong & None & Avg. & Max & Avg. & Min & Max  & Easy & Medium & Hard \\
\midrule
\Textonly   & 83.6\%  & 46   & 9   & 1   & 1.83 & 12  & 5:19  & 0:43  & 27:24 & 23   & 20     & 13   \\
\addlinespace[1.5pt]

\Multimodal & 85.7\%  & 48   & 5   & 2   & 1.22 & 14  & 4:23  & 0:26  & 24:14 & 32   & 16     & 7    \\
\addlinespace[1.5pt]

All         & 84.7\%  & 94   & 14  & 3   & 1.50 & 14  & 4:46  & 0:26  & 27:24 & 55   & 36     & 20  \\
\bottomrule
\end{tabular}
}
\caption{Humans achieve 84.7\% accuracy on \name\ and were generally able to find answers smoothly (1.5 dead ends on average) and quickly (4 min 46 sec). The \emph{None} label marks questions abandoned after 15 minutes. \emph{Perceived difficulty} reflects annotators' subjective assessment of question difficulty. The minimum number of dead ends was zero. 
% Humans achieved 84.7\% accuracy and were generally able to find answers smoothly (1.5 dead end) and quickly (4 min 46 sec). 
% \ys{mention what is perceived diff}
}
\label{tab:human_performance}
\end{table}
% https://docs.google.com/spreadsheets/d/1xuJAWKn7tpO-H0VnHxomCtW36tookVQcgoZpjBjHxVk/edit?usp=sharing

\section{Human error examples}\label{appendix:human_error_examples}

\tableref{tab:human_errors} lists six key causes of human errors mentioned in \sectionref{sec:human_eval_results}, along with corresponding examples and explanations. 

\begin{table}[h]
\fontsize{9}{11}\selectfont
\centering
\resizebox{0.97\columnwidth}{!}{%
\begin{tabular}{@{}p{9cm}p{6cm}}

\toprule
\textbf{\fontsize{10}{12}\selectfont Error type \& question}  &  \textbf{\fontsize{10}{12}\selectfont Explanation} \\
\midrule

\textbf{Type}: \colorbox{blue!20}{Missing details in the question or answer}; \colorbox{blue!20}{Lack of topic} \colorbox{blue!20}{knowledge};  \colorbox{blue!20}{Suboptimal source selection}.

\textbf{Question}: Which bird(s) in the Wingspan bird cards from the Asia expansion have the word ``Great'' (but not ``Greater'') in their name and can reside exclusively in wetlands?  & 

The annotator (1) was unfamiliar with the game, (2) found a Wingspan card website lacking expansion and habitat details, then identified and searched bird with ``great'' individually, and (3) overlooked the Asia expansion requirement.
\\\midrule

\textbf{Type}: \colorbox{blue!20}{Obvious oversight}

\textbf{Questions}: What does the October 2024 edition (817) of Tinkle magazine teach readers to make on page 36?  &

The annotator accessed a relevant magazine page but overlooked the answer on it.
\\\midrule

\textbf{Type}: \colorbox{blue!20}{Complexity of task}

\textbf{Question}: On February 10, 2016, at 2:00 AM, a tagged Altai Snow Leopard was observed in West Mongolia, according to the data from USGS. What was the straight-line distance it traveled by the time it was observed again at 6:00 AM on the same day? &

The annotator easily found the website but gave up after failing to identify the correct map markers representing the Snow Leopard’s location at the specified date and time.
\\\bottomrule

\end{tabular}
}
\caption{Examples of human errors. The annotators answered 14 questions incorrectly and gave up on 3, primarily due to five key reasons listed in the table. We provide three examples, each with attributed error reasons and detailed explanation. The two more frequent error reasons are \colorbox{blue!20}{missing details in the question or answer} and \colorbox{blue!20}{lack of topic knowledge}.
% Example questions and explanations are provided.
}
\label{tab:human_errors}
\end{table}

% Reasons for Incorrect or Abandoned Answers
% x Incorrect source/tool selection (12, 37, 64)
% x Missed details in the question or answer (16, 21, 23, 28, 71, 101; 78)
% x Lack of topic understanding (18, 32, 68, 104)
% x Obvious oversight (87)

% Reasons for Giving Up
% x Task too complex (45)
% x Lack of topic understanding (49)
% x Conventional way of thinking (91)

\section{Details of automatic agent answer evaluator}\label{appendix:auto_rater}

This section provides the implementation details of the automatic evaluator in \sectionref{sec:agent_setup}.

\noindent\textbf{Prompt of the automatic evaluator} An example of the prompt used for automatic evaluation is shown in \tableref{tab:auto_rater_prompt}. Because the prompt contains 17 in-context examples, the full text is omitted here. The complete prompt and the code of the evaluator are available at \url{https://bear-cubs.github.io/}.

\begin{table}[t]
  \centering
  \small
  \begin{tabular}{@{}p{0.98\linewidth}@{}}
    \toprule
    \textbf{Prompt of the automatic agent response evaluator}\\

    \addlinespace[1ex]
    
    Given a question, its gold answer, and an AI generated answer, your task is to check for the following four cases:\\

    \addlinespace[1ex]
    
    1.~no\_answer: AI generated answer is empty.
    
    2.~no\_direct\_answer: Agent abstains or hedges with modal verbs (e.g., “may”, “could be”) or phrases expressing uncertainty such as “according to common sense” or “I am leaning toward.”
    
    3.~correct: AI returns a direct and unambiguous answer which entails the gold answer.\\[1ex]
    4.~wrong: AI returns a direct and unambiguous answer which, however, does not entail the gold answer. \\

    \addlinespace[1ex]
    
    Every question has an answer. If the AI answer rejects the premise of a question, it should be labeled as ``wrong.''\\

    \addlinespace[1ex]
    
    The gold answers are always concise and never presented as a list. They may be a number, a price, a word, or a phrase, among other types. In contrast, AI answers can be long, such as a paragraph, with an answer embedded within. The embedded answer may be correct but differ in format (e.g., how a number is written). Nonetheless, they should express the same information (i.e., 3 and four are different, 303kg and 302kg are different.) Below, I provide examples of how to decide which of the four labels to assign to the AI answer along with explanations. However, in your evaluation, you do not need to explain your decisions.\\

    \addlinespace[1ex]
    
    \textbf{Example 1:}
    
    Question: What is the name of the bird(s) in the Wingspan bird cards from the Asia expansion have the word "Great" (but not "Greater") in their name and can reside exclusively in wetlands?
    
    Gold Answer: Great Cormorant
    
    AI Answer:
    
    Label: no\_answer
    
    Explanation: The AI answer is empty. Hence, the label is ``no\_answer.''\\

    \addlinespace[1ex]
    
    \textbf{Example 2:}
    
    Question: In one episode of Wissen Macht Ah! in April 2023, the hosts demonstrated how to make a DIY water filter in the wild. What was the second filtering material that they placed into the plastic bottle? Answer the question with the exact word used in the episode?
    
    Gold Answer: Activkohle
    
    AI Answer: Sorry, I don't have access to specific episodes of *Wissen Macht Ah!* or their content, including the exact word used in the April 2023 episode. You might want to check the episode directly or refer to its official resources for the precise information.
    
    Label: no\_direct\_answer
    
    Explanation: The AI answer does not provide a direct answer to the question and instead states that it does not have access to the specific episode's content.\\

    \addlinespace[1ex]
    
    \multicolumn{1}{@{}p{0.98\linewidth}@{}}{%
    \centering\textit{Further examples omitted for the sake of space.}%
    }\\

    \addlinespace[1ex]
    
    Now it is your turn to label the following AI answers. You do not need to explain your decisions.\\

    \addlinespace[1ex]

    Question: \{question\}
    
    Gold Answer: \{gold\_answer\}
    
    AI Answer: \{ai\_output\}
    
    Label:\\
    
    \bottomrule
  \end{tabular}
  \caption{Example of the prompt used for the automatic agent response evaluator. Examples 3 to 17 are omitted for the sake of space.}
  \label{tab:auto_rater_prompt}
\end{table}

\noindent\textbf{Modal and hyperparameters} The evaluator is built on \texttt{gpt-4o-2024-11-20} as its base LLM, with the temperature fixed at 0. Running the evaluator on 111 questions takes approximately 1 minute and 30 seconds, costing around \$0.80 USD.

\noindent\textbf{Evaluator performance} As shown in the prompt in \tableref{tab:auto_rater_prompt}, the evaluator performs a four-way classification task. We also report results with the four labels collapsed into two categories, \texttt{correct} and \texttt{incorrect}. The evaluator was applied to the outputs of five agents: Grok3 DeepSearch, OpenAI Deep Research, Anthropic Computer Use, Convergence AI Proxy, and OpenAI Operator. The results for both the four-way and binary settings are presented in \tableref{tab:auto_rater_accuracy}. To verify the reliability of the evaluator, we ran it three times on outputs from Anthropic and Proxy, obtaining identical results in all runs.

\begin{table}[t]
  \centering
  \small
  \begin{tabular}{@{}lcc@{}}
    \toprule
    \textbf{Agent} & \textbf{4-Way Accuracy} & \textbf{Binary Accuracy} \\
    \midrule
    Grok3 Deep Search      & 99.1\% & 99.1\% \\
    OpenAI Deep Research   & 96.4\% & 96.4\% \\
    Anthropic Computer Use & 99.1\% & 100\% \\
    Convergence AI Proxy   & 99.1\% & 100\% \\
    OpenAI Operator        & 97.3\% & 98.2\% \\
    
    \bottomrule
  \end{tabular}
  \caption{Evaluator accuracy by agent in the four-way classification setting. Binary accuracy is derived by collapsing the four labels into two categories: \texttt{correct} and \texttt{incorrect}.}
  \label{tab:auto_rater_accuracy}
\end{table}

\section{Detailed agent performance results}\label{appendix:full_agent_results}

\tableref{tab:model_performance_full} provides detailed agent results, grouped by all questions, \textonly, and \multimodal~questions.

\begin{table*}[h]
\fontsize{11}{13}\selectfont
\centering
\resizebox{0.96\textwidth}{!}{%
\begin{tabular}{llp{1.1cm}p{1cm}p{1.5cm}p{1cm}llllll}
\toprule
\multirow{2}{*}{\textbf{Model}} &
  \multirow{2}{*}{\textbf{Accuracy}} &
  \multicolumn{4}{c}{\textbf{Answer label}} &
  \multicolumn{3}{c}{\textbf{Average time}} &
  \multicolumn{3}{c}{\textbf{Correct answer source attribution}} \\
\cmidrule(lr){3-6}
\cmidrule(lr){7-9}
\cmidrule(lr){10-12}
  &
  & 
  Correct &
  Wrong &
  Uncertain&
  None &
  Correct &
  Wrong   &
  Uncertain &
  Ungrounded &
  Primary &
  Secondary \\
\midrule
\addlinespace[5pt]
\multicolumn{12}{c}{\textbf{All Questions}}                                     \\
\cmidrule{1-12}
\addlinespace[5pt]
\rowcolor{yellow!30}
GPT-4o zero-shot& 2.7\%  & 3   & 53  & 55  & 0  & --- & --- & --- & ---    & ---    & ---  \\
\rowcolor{yellow!30}
DeepSeek R1 zero-shot
             & 8.1\%  & 9   & 82  & 19  & 1  & --- & --- & --- & ---    & ---    & ---  \\
\rowcolor{yellow!30}
GPT-4o + Google Search & 0.0\%  & 0   & 4   & 0   & 107& --- & --- & --- & 0      & 0      & 0   \\
\rowcolor{yellow!30}
DeepSeek R1 + Google Search
             & 1.8\%  & 2   & 16  & 0  & 93  & ---    & ---    & ---  & 1   & 0   & 1    \\
\rowcolor{yellow!30}
Perplexity \texttt{sonar-pro}
             & 5.4\%  & 6   & 30   & 58   & 17 & ---  & ---  & ---  & 2    & 4   & 0  \\
\rowcolor{blue!5}
Grok3 DeepSearch       & 11.7\% & 13  & 95  & 2   & 1  & 1:09   & 1:24   & 2:05  & 3   & 8   & 2   \\
\rowcolor{blue!5}
OpenAI Deep Research& 36.0\% & 40  & 69  & 1   & 1  & 4:37	  & 9:00   & 3:58 & 5   & 25  & 10  \\
\rowcolor{blue!5}
Google Deep Research& 23.4\% & 26  & 39  & 46   & 0  & 4:21	  & 4:00   & 4:39 & 1   & 25  & 0  \\
Convergence AI Proxy        & 12.6\% & 14  & 44  & 34  & 19 & 1:52	  & 2:41   & 5:24 & 0   & 13  & 1   \\
Anthropic Computer Use & 14.4\% & 16  & 22  & 73  & 0  & 2:24	  & 2:35   & 3:35 & 0   & 14  & 2   \\
OpenAI Operator     & 23.4\% & 26  & 43  & 13  & 29 & 2:59	  & 3:58   & 8:06 & 1   & 24  & 1   \\
OpenAI ChatGPT Agent     & 65.8\%\% & 73  & 30  & 3  & 5 & 9:16	  & 16:14   & 25:55 & 1   & 72  & 0   \\
\rowcolor{green!15}
Human        & 84.7\% & 94  & 14  & --- & 3  & 4:24   & 5:44   & ---  & --- & --- & --- \\
\addlinespace[5pt]
\cmidrule{1-12}
\multicolumn{12}{c}{\textbf{\textcolor{bt}{Text-based Questions}}}  \\
\cmidrule{1-12}
\addlinespace[5pt]
\rowcolor{yellow!30}
GPT-4o zero-shot& 5.4\%  & 3   & 30  & 23  & 0  & --- & --- & --- & ---  & ---  & --- \\
\rowcolor{yellow!30}
DeepSeek R1 zero-shot
             & 10.7\% & 6   & 38  & 12  & 0  & --- & --- & --- & ---  & ---  & --- \\
\rowcolor{yellow!30}
GPT-4o + Google Search & 0.0\%  & 0   & 4   & 0   & 52 & ---  & ---  & ---  & 0   & 0   & 0   \\
\rowcolor{yellow!30}
DeepSeek R1 + Google Search 
             & 3.6\%  & 2   & 8   & 0   & 46 & ---  & ---  & ---  & 1   & 0   & 1   \\
\rowcolor{yellow!30}
Perplexity \texttt{sonar-pro}
             & 8.9\%  & 5   & 19   & 26   & 6 & ---  & ---  & ---  & 1    & 4   & 0  \\
\rowcolor{blue!5}
Grok3 DeepSearch       & 21.4\% & 12  & 42  & 2   & 0  & 1:07 &1:30  & ---  & 2   & 8   & 2   \\
\rowcolor{blue!5}
OpenAI Deep Research& 60.7\% & 34  & 21  & 0   & 1  & 4:12 & 8:32 & ---  & 0   & 24  & 10  \\
\rowcolor{blue!5}
Google Deep Research& 42.9\% & 24  & 13  & 19   & 0  & 4:22 & 4:13 & 5:13  & 0   & 24  & 0  \\
Convergence AI Proxy        & 16.1\% & 9   & 26  & 14  & 7  & 2:05 & 2:36 & 4:11 & 0   & 8   & 1   \\
Anthropic Computer Use & 19.6\% & 11  & 12  & 33  & 0  & 2:49 & 2:27 & 3:21 & 0   & 9   & 2   \\
OpenAI Operator     & 33.9\% & 19  & 24  & 3   & 10 & 3:13 & 3:51 & 7:24 & 0   & 18  & 1   \\
OpenAI ChatGPT Agent     & 76.8\%\% & 43  & 10  & 1  & 2 & 6:29	  & 12:06   & 43:49 & 0   & 43  & 0   \\
\rowcolor{green!15}
Human        & 83.6\% & 46  & 8   & --- & 1  & 5:09 & 5:07 & ---  & --- & --- & --- \\
\addlinespace[5pt]
\cmidrule{1-12}
\multicolumn{12}{c}{\textbf{\textcolor{yt}{Multimodal Questions}}}                                                         \\
\cmidrule{1-12}
\addlinespace[5pt]
\rowcolor{yellow!30}
GPT-4o zero-shot& 0.0\%  & 0   & 23  & 32  & 0  & ---  & ---  & ---  & ---  & --- & --- \\
\rowcolor{yellow!30}
DeepSeek R1 zero-shot
             & 5.5\%  & 3   & 44  & 7   & 1  & ---  & ---  & ---  & ---  & --- & --- \\
\rowcolor{yellow!30}
GPT-4o + Google Search & 0.0\%  & 0   & 0   & 0   & 55 & ---  & ---  & ---  & 0    & 0   & 0  \\
\rowcolor{yellow!30}
DeepSeek R1 + Google Search
             & 0.0\%  & 0   & 8   & 0   & 47 & ---  & ---  & ---  & 0    & 0   & 0  \\
\rowcolor{yellow!30}
Perplexity \texttt{sonar-pro}
             & 1.8\%  & 1   & 11   & 32   & 11 & ---  & ---  & ---  & 1    & 0   & 0  \\
\rowcolor{blue!5}
Grok3 DeepSearch       & 1.8\%  & 1   & 53  & 0   & 1  & 1:25	& 1:20 & ---  & 1    & 0   & 0  \\
\rowcolor{blue!5}
OpenAI Deep Research& 10.9\%  & 6   & 48  & 1   & 0  & 6:58	& 9:12 & 3:58 & 5    & 1   & 0  \\
\rowcolor{blue!5}
Google Deep Research& 3.6\%  & 2   & 26  & 27   & 0  & 4:15	& 3:53 & 4:16 & 1    & 1   & 0  \\
Convergence AI Proxy        & 9.1\%  & 5   & 18  & 20  & 12 & 1:29 & 2:48 & 10:42& 0    & 5   & 0  \\
Anthropic Computer Use & 9.1\%  & 5   & 10  & 40  & 0  & 1:29	& 2:45 & 3:47 & 0    & 5   & 0  \\
OpenAI Operator     & 12.7\% & 7   & 19  & 10  & 19 & 2:21	& 4:06 & 8:19 & 1    & 6   & 0  \\
OpenAI ChatGPT Agent     & 54.5\%\% & 30  & 20  & 2  & 3 & 13:15	  & 18:19   & 16.58 & 1   & 29  & 0   \\
\rowcolor{green!15}
Human        & 85.7\% & 48  & 6   & --- & 2  & 3:41	& 6:41 & ---  & ---  & --- & --- \\
\bottomrule
\end{tabular}
}
\caption{Model performance on \name~with the \colorbox{yellow!30}{baselines} \texttt{gpt-4o-2024-11-20} and DeepSeek R1 (with/without Google Search), Perplexity \texttt{sonar-pro}, \colorbox{blue!5}{deep research agents}, computer-use agents, and \colorbox{green!15}{human performance}. `Uncertain' indicates that the agent did not return a concrete answer (e.g., abstention), while `None' means either a baseline model returned ``No answer found'' or an agent entered a dead loop and failed to provide any response. `Ungrounded' answers are those not based on a source but rather on the agent's internal knowledge or reasoning. `Primary' denotes an answer derived from a reliable source or the source specified in the question, whereas `Secondary' refers to an answer obtained from an unreliable source or a source not mentioned in the question.}
\label{tab:model_performance_full}
\end{table*}

\section{Justifying size of \name}\label{appendix:std_on_bearcubs}

\textbf{Small size benchmarks are not rare.} \name\ currently contains 111 questions. Such a benchmark size is not uncommon in prior popular benchmarks. The widely used AIME dataset \citep{AIME2024} contains only 30 questions and is still adopted in both evaluation and model development efforts such as by OpenAI \citep{openai2025o3o4mini} and DeepSeek \citep{deepseekai2025deepseekv3technicalreport}. Similarly, CodeElo \citep{quan2025codeelo} includes just 387 problems, and the Proof or Bluff dataset \citep{petrov2025proofbluffevaluatingllms} uses only six USAMO questions for testing.

\textbf{Creating high-quality \name\ questions is non-trivial.} In our experience, freelance annotators were only able to write a small number of accepted questions before quality declined noticeably. Additionally, OpenAI’s Deep Research agent frequently identified unintended shortcuts, especially in multimodal setting, which led us to discard many questions to preserve benchmark integrity.

\textbf{\name\ remains a meaningful benchmark.} This is due to two key factors. First, the search space for \name\ questions spans the entire real-world web, unlike, for example, constrained multiple-choice formats. This makes guessing the correct answer difficult and reduces accuracy variance. Second, solving even a single \name\ question typically requires a complex mix of retrieval, reasoning, and synthesis, making each example informative for evaluating model capabilities. To support the first claim, we evaluated model accuracy variance on \name\ by running DeepSeek R1 with and without Serper across four independent trials. The results, summarized below, show minimal standard deviation in accuracy across runs.\footnote{The first run was conducted in March 2025 using the Fireworks AI API; subsequent runs were carried out via OpenRouter.}

\begin{table}[h]
\centering
\begin{tabular}{@{}lrrrrr@{}}
\toprule
\textbf{Setup} & 1st Run & 2nd Run & 3rd Run & 4th Run & Std. Dev. \\
\midrule
DeepSeek R1 w/o Serper & 7.2\% & 5.4\% & 5.4\% & 5.4\% & 0.8\% \\
DeepSeek R1 w/ Serper  & 1.8\% & 0.9\% & 2.7\% & 2.7\% & 0.7\% \\
\bottomrule
\end{tabular}
\caption{Accuracy standard deviation across four trials of DeepSeek R1 on \name, with and without Serper. Results show low standard deviation, indicating stable performance.}
\label{tab:std_deepseek}
\end{table}

\section{Model correct answer source proportions}\label{appendix:correct_answer_source_prop}

\begin{figure}[h]
\begin{center}
\includegraphics[scale=0.2]{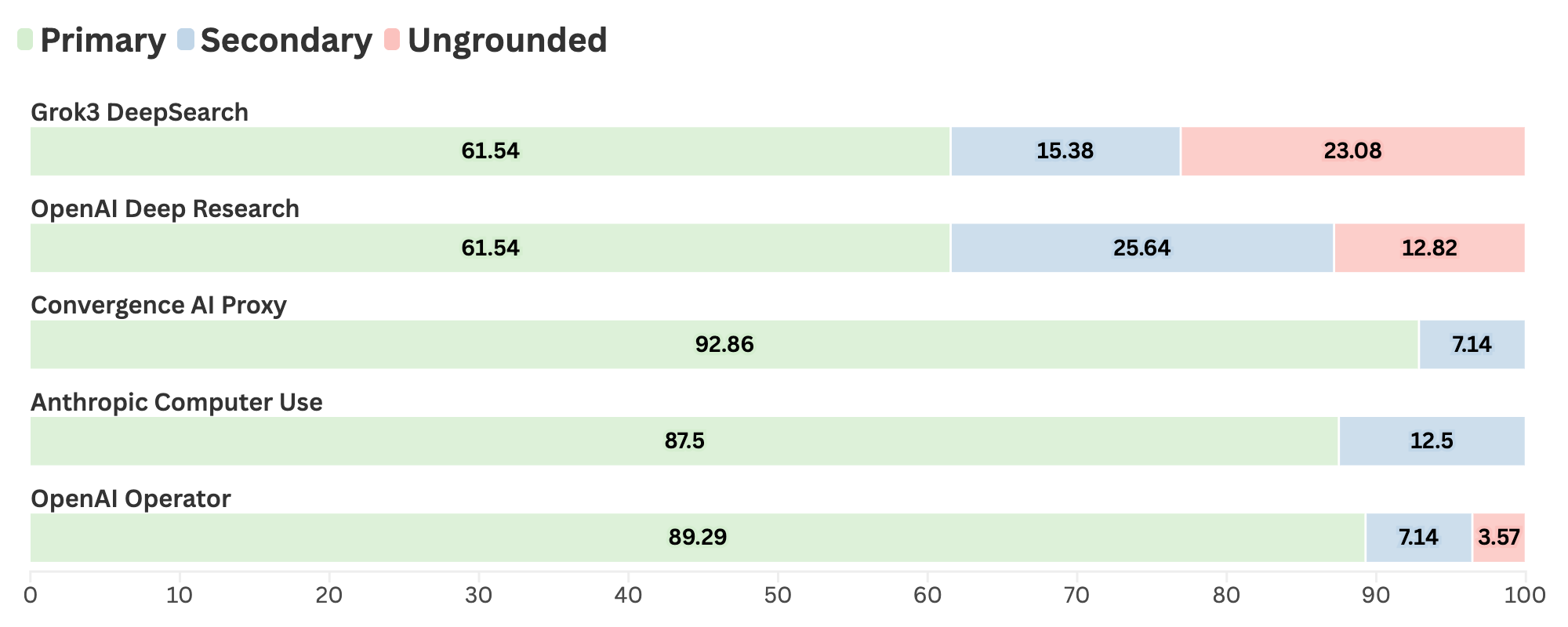}
\end{center}
\caption{Proportion of correct answers within each agent categorized by reliance on primary sources, secondary sources, and ungrounded responses.}
\label{fig:groundedness}
\end{figure}

\figureref{fig:groundedness} presents the proportion of correct answers within each agent categorized by reliance on primary
sources, secondary sources, and ungrounded responses. The analysis is provided in \sectionref{sec:discussion}.

\section{Agent-specific behavior}\label{appendix:agent_specific_behavior}

We continue the discussion in \sectionref{sec:discussion} and present interesting agent-specific behavior and the challenges those agents face that hinder their utility.

We observe that Grok 3 returns its search and reasoning trajectory in a non-English language if it appears in the input, which can be unhelpful for users seeking assistance for that language. Grok 3 also never abstains, always generating a response (see \tableref{tab:model_performance_all_only})---a behavior that needs user study to determine its desirability. Meanwhile, Computer Use sometimes deems a task impossible without attempting it, despite users expecting an effort and justification. These issues, along with the broader challenges discussed above, highlight the need for improved adaptability, transparency, and user-centered refinement in LLM-based computer use agents.
\end{document}